%% file: camera_ready.tex
\documentclass[10pt,twocolumn,letterpaper]{article}

\usepackage{cvpr}      % To produce the REVIEW version
\input{preamble}

\definecolor{cvprblue}{rgb}{0.21,0.49,0.74}
\usepackage[pagebackref,breaklinks,colorlinks,allcolors=cvprblue]{hyperref}

\usepackage{caption}  % for \captionof

\usepackage{blindtext}% for dummy text only
\usepackage{graphicx}
\usepackage{tikz}
\usetikzlibrary{positioning}
\usepackage{pgfplots}
\usepackage{bm}
\usepackage{multirow}
\usepackage[table]{xcolor}
\usepackage{makecell}
\usepackage{calc}
\usepackage{algorithm}
\usepackage{algorithmic}
\usepackage{pifont}% http://ctan.org/pkg/pifont
\newcommand{\xmark}{\ding{55}}%
\newcommand{\std}[1]{{\scriptsize$\pm$ #1}}

\def\paperID{6204} % *** Enter the Paper ID here
\def\confName{CVPR}
\def\confYear{2026}

\title{Continual Distillation of Teachers from Different Domains}

\author{
Nicolas Michel$^{1,4}$ \quad
Maorong Wang$^{2}$ \quad
Jiangpeng He$^{3, \dagger}$ \quad
Toshihiko Yamasaki$^{1, \dagger}$\vspace{0.3em} \\
$^{1}$The University of Tokyo \quad
$^{2}$National Institute of Informatics \\
$^{3}$Indiana University Bloomington \quad
$^{4}$Japanese-French Laboratory for Informatics, CNRS \vspace{0.3em} \\
{\tt\small \{nicolas, yamasaki\}@cvm.t.u-tokyo.ac.jp,} {\tt\small maorong@nii.ac.jp, jhe2@iu.edu}
}

\begin{document}

\twocolumn[{%
\renewcommand\twocolumn[1][]{#1}%
\maketitle

\def\thefootnote{$\dagger$}\footnotetext{Equal supervision.}
\def\thefootnote{\arabic{footnote}} % Resets footnote counter for the rest of the paper

}]

\begin{figure*}[t]
\centering
\includegraphics[width=0.8\linewidth]{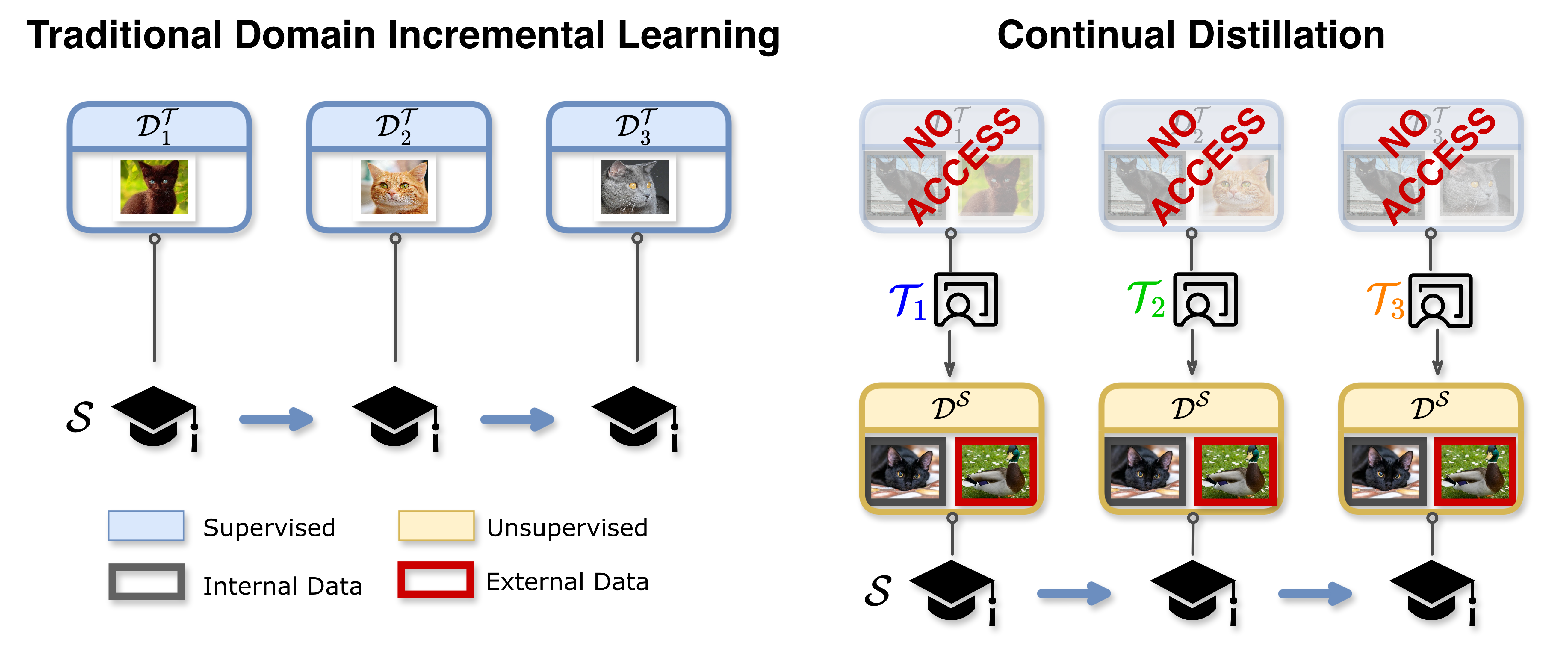}
\vspace{-4pt}
\caption{Overview of the Continual Distillation problem. A student model $\mathcal{S}$ learns through distillation from a sequence of teachers $\{\mathcal{T}_1, \mathcal{T}_2, \mathcal{T}_3\}$. During distillation, part of the teachers' training data is unavailable. The objective is for the student to maintain high performance on \textbf{all domains known to the teachers}, but not necessarily introduced to the student.}
\vspace{-4pt}
\label{fig:overview}
\end{figure*}

\begin{abstract}

% CVPR version%%%%%%%%% ABSTRACT
\def\thefootnote{$\dagger$}\footnotetext{Equal supervision.}
\def\thefootnote{\arabic{footnote}} % Resets footnote counter for the rest of the paper
Deep learning models continue to scale, with some requiring more storage than many large-scale datasets. Thus, we introduce a new paradigm: Continual Distillation (CD), where a student learns sequentially from a stream of teacher models without retaining access to earlier teachers. CD faces two challenges: teacher training data is unavailable, and teachers have varying expertise. We show that external unlabeled data enables Unseen Knowledge Transfer (UKT), allowing the student to acquire information from domains not present in the training data, while known to the teacher. We also show that sequential distillation causes Unseen Knowledge Forgetting (UKF) when transferred knowledge is lost after training on later teachers. To better trade off between UKT and UKF, we propose Self External Data Distillation (SE2D), a method that preserves logits on external data to stabilize learning across heterogeneous teachers. Experiments on multiple benchmarks show that SE2D reduces UKF and improves cross-domain generalization. The code and implementation for this work are publicly available at: \url{https://github.com/Nicolas1203/continual_distillation}.
\end{abstract}
\vspace{-10pt}

\section{Introduction}
\label{sec:intro}
%% CVPR version
Over the last decade, deep learning models have reached unprecedented scales, creating a growing need for computation-efficient strategies. Consequently, Continual Learning (CL)~\cite{de_lange_continual_2021, parisi2019continual, wang_comprehensive_2023} has become a major branch of contemporary deep learning research. The main idea is straightforward: a model is trained on a sequence of datasets where previous data becomes unavailable over time. The rationale is that re-training on both past and current data as new data arrives can be computationally expensive or require substantial storage. In this context, data access is limited during training.

With the recent adoption of Foundation Models (FMs)~\cite{devlin2019bert, liu2019roberta, dosovitskiy2020vit, radford2021learning, achiam2023gpt} as the backbone of modern deep learning, we propose a new paradigm, \textbf{Continual Distillation (CD)}, tailored to FMs, illustrated in Figure~\ref{fig:overview}. In CD, instead of learning from a sequence of datasets, we propose to \textit{learn from a sequence of models trained on different datasets}. Specifically, a single student model learns sequentially from multiple teacher models without retaining access to earlier ones. Similar to training data in CL, new FMs are regularly introduced over time, and storing them is cumbersome since they can rapidly require more storage than large-scale datasets. For instance, storing 10B parameters requires approximately 38GB~\cite{PyTorch_2024}, while FMs often exceed 100B parameters. Even accessing FMs through restricted APIs presents serious limitations, as previous versions may become unavailable after updates. Our CD setup reflects a realistic scenario in which one aims to leverage the ever-evolving stream of FMs to train smaller and more specialized models through distillation. In analogy to CL, where a single model benefits from a continuous stream of data, in CD, a single student model benefits from a continuous stream of teacher models.

However, CD introduces two main challenges. First, the original training data of a foundation model is typically unavailable, undisclosed, or prohibitively large to reuse~\cite{awais2025foundation}. Thus, choosing distillation data is critical, and such data can potentially emerge from a domain unknown to the teacher~\cite{wang2024data_free_kd_dist_shift}. Second, each teacher generally exhibits distinct abilities and performance; for instance, one may excel at recognizing animals, while another may specialize in distinguishing insects.

In this work, we focus on the case of Continual Distillation, where, analogous to Domain Incremental Learning, teachers have been trained on a domain incremental set of datasets. However, we assume that teachers share partially overlapping domains. We believe this assumption to be realistic, as, for example, it is safe to assume that all FMs have been trained on ImageNet. Therefore, we make the following key assumptions: (1) each teacher is trained on data from different domains while sharing a specific domain; (2) the training data for distillation is fixed; and (3) the training data is unlabeled. The ultimate objective is to obtain a student model that achieves high performance on every domain known by at least one teacher, without having access to all teacher training data or labels. In this context, we find that training data can be decomposed into two categories: \textbf{External Data (ED)}, unknown to all teachers, and \textbf{Internal Data (ID)}, known to all teachers. Importantly, we discover that ED enables the transfer of knowledge from domains unseen by the student but known by the teacher during distillation, which we refer to as \textbf{Unseen Knowledge Transfer (UKT)}. Naturally, we observe that the student inevitably forgets unseen knowledge transferred by previous teachers when learning from future ones. We refer to this phenomenon as \textbf{Unseen Knowledge Forgetting (UKF)}. The central problem of CD becomes reaching an optimal UKT-UKF trade-off.

Experimentally, we observe that while the usage of ED enables UKT, mainstream distillation strategies fail to address UKF, as their primary focus relies on maximizing knowledge transfer only. We therefore propose Self External Data Distillation (SE2D), a CL-inspired method tailored for the CD problem, focusing on preserving logits of ED to maintain performance on domains unseen by the student. SE2D allows positive UKT while maintaining performance in older domains. Our contributions are as follows:
\begin{itemize}
\item We introduce the paradigm of Continual Distillation, motivated by the practical challenges that arise when previous teacher models are no longer accessible.
\item We demonstrate that the choice of distillation data is crucial for transferring knowledge to domains unseen by the student, and choose to take advantage of External Data, unknown to the teachers.
\item We identify Unseen Knowledge Forgetting, mitigate it by preserving External Data logits, and validate our approach across various benchmarks.
\end{itemize}

\section{Related Work}
\label{sec:related}
% \vspace{-4pt}

\textbf{Continual Learning.}
Continual Learning (CL)~\cite{chen2018lifelong, parisi2019continual} focuses on enabling models to learn from a sequence of tasks while retaining previously acquired knowledge. Traditionally, each CL task is defined by different non-overlapping datasets with unique properties. In Class Incremental Learning (CIL)~\cite{dong_class-incremental_2023}, each task is composed of a unique set of classes. In Domain Incremental Learning (DIL)~\cite{buzzega_dark_2020}, classes are shared, but input distributions vary. While access and change in data have been widely studied, we are, to the best of our knowledge, the first to consider an alternative scenario where data is fixed but accessible model change over time. In this case, we focus on a situation where the teachers' training domain differs, not unlike a DIL setup, where each teacher would be trained on a specific task of a DIL setting. Since foundation models become ubiquitous~\cite{radford2021learning}, large, and ever-changing, a paradigm shift from ever-evolving data to ever-evolving teachers appears relevant. While Knowledge Distillation is often employed in CL to mitigate forgetting~\cite{michel2024rethinking, rebuffi2017icarl, touvron2023llama}, we instead investigate the phenomenon of forgetting within distillation itself, treating it as the central challenge of our study.

\noindent\textbf{Knowledge Distillation.}
% Maorong
Knowledge Distillation (KD)~\cite{hinton_distilling_2015} is a technique that allows a smaller student model to replicate the behavior of a larger, more capable teacher model. Ideally, KD assumes that both the teacher and student are trained on the same dataset, and has been widely used for model compression~\cite{romero2014fitnets, zhao2022decoupled} and transfer learning~\cite{yim2017gift}. When such data is not available, distillation is considered to be data-free~\cite{wang2024data_free_kd_dist_shift}. In that context, the objective is usually to reproduce the teacher's training domain as closely as possible.
Thanks to the capability of knowledge transfer from KD, it has become one of the cornerstones for solving the forgetting problem in CL~\cite{rebuffi2017icarl,douillard2020podnet}. Nevertheless, the ideal setting (i.e., the students are distilled with the original teachers' training dataset) of KD rarely holds in CL due to the unavailability of historical data. Consequently, CL methods conduct distillation with new training samples~\cite{li2017learning}, small memory buffer samples~\cite{michel2024rethinking}, generated samples~\cite{wu2018memory}, or adversarial samples~\cite{goswami2024resurrecting}. 

While prior work has explored multi-teacher knowledge distillation and the use of distillation in continual learning, our setting differs in several key aspects. 
First, existing multi-teacher distillation methods typically assume simultaneous access to all teachers, whereas we consider a sequential setting where only one teacher is accessible at a time. 
Second, unlike continual learning approaches that focus on data streams, our formulation assumes a fixed dataset and instead models a stream of evolving teacher models. 
Finally, in contrast to standard distillation-based continual learning methods, we do not assume access to past data or replay buffers, nor to previously seen teachers. 
This combination of constraints defines a distinct and practically relevant scenario that, to the best of our knowledge, has not been explicitly studied before.
We therefore frame this problem as Continual Distillation, emphasizing the shift from data-centric to model-centric CL.

% Original
% Knowledge Distillation (KD)~\cite{hinton_distilling_2015} is a technique that allows a smaller student model to replicate the behavior of a larger, more capable teacher model. Traditionally, KD assumes that both the teacher and student are trained on the same dataset, and has been widely used for model compression~\cite{romero2014fitnets, zhao2022decoupled} and transfer learning~\cite{yim2017gift}.
% With the development of the KD  community, Data-Free KD (DFKD)~\cite{wang2024data_free_kd_dist_shift} has emerged as a crucial technique for scenarios where the teacher’s original training data is unavailable. Representative DFKD techniques include deep inversion, generative model-assisted sampling, or using out-of-domain data to approximate the teacher’s behavior. While effective, these approaches often introduce distributional shifts between the distillation data and the teacher’s original training data, which can negatively affect the quality of the distilled knowledge.
\begin{figure}[t]
    \centering
    \includegraphics[width=0.8\linewidth]{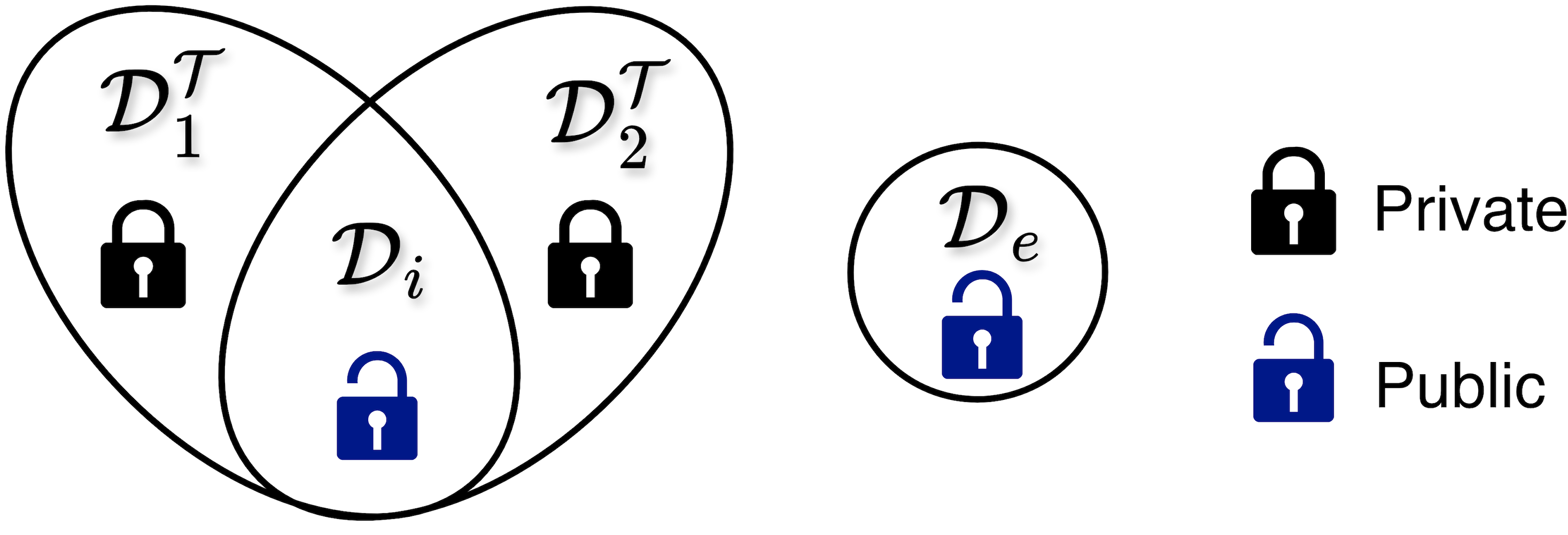}
    \caption{Visualization $\mathcal{D}_e$ and $\mathcal{D}_i$ and different teacher training domains in our experimental setting. $\mathcal{D}_i$ and $\mathcal{D}_e$ are publicly available like ImageNet or Wikipedia. However, some private or restricted data might be included in training, such as medical data.}
    \label{fig:problem_scenario}
    \vspace{-10pt}
\end{figure}

\section{Continual Distillation}
\label{sec:method}
\subsection{Generic Definition}
We define Continual Distillation (CD) as the process of distilling the knowledge from a \textbf{sequence of teacher models} continuously into \textbf{one student model}, on a \textbf{fixed dataset}. When distilling from one teacher to the student, other teachers are considered unavailable. The distillation process from a given teacher to the student is analogous to a task in standard CL. Formally, given a sequence of teachers $\{\mathcal{T}_0, \mathcal{T}_1, \dots, \mathcal{T}_N\}$, each trained on a dataset $\mathcal{D}_t^\mathcal{T}$, the student $\mathcal{S}$ is optimized to minimize distillation loss $\mathcal{L}_{dist}$, with respect to $\mathcal{T}_t$ on a distillation dataset $\mathcal{D}^\mathcal{S}$. Importantly, we only consider distillation and no label-dependent loss is considered. In this work, we focus on logits-based distillation as representations are architecture-dependent, computation-intensive, and require access to the entire teacher model. We present the overall procedure in Figure~\ref{fig:overview}. We denote by $\mathbb{D}_\mathcal{S}$($\mathcal{T}_{t}$, $\mathcal{D}^\mathcal{S})$ the operation of distilling from teacher $\mathcal{T}$ trained with $\mathcal{D}^\mathcal{T}_{t}$ to student $\mathcal{S}$ on distillation dataset $\mathcal{D}^\mathcal{S}$.

\subsection{Specific Problem Scenario}
\label{sec:scenarios}

Traditional KD assumes that the teacher training datasets are available for distillation. In CD, not only are such datasets considered unavailable, but dataset domains might differ from one teacher training to another. In other words, $\mathcal{D}_t^\mathcal{T}$, $\mathcal{D}_{t^\prime}^\mathcal{T}$ and $\mathcal{D}^\mathcal{S}$ may cover partially or totally different domains. Therefore, various scenarios can be defined in CD depending on the domain overlap between teachers training data $\mathcal{D}_t^\mathcal{T}$ and distillation data $\mathcal{D}^\mathcal{S}$. Realistically, when considering FMs, it is safe to assume that part of the training data is shared. Typically, publicly available datasets such as Wikipedia or ImageNet are commonly used for training FMs. However, additional data, exclusive to a specific model, might also be included during training. A visualisation of such a scenario is presented in Figure~\ref{fig:problem_scenario}. Formally, we consider the case of partially exclusive teacher domains such that all teachers share a specific domain:
\begin{figure}[t]
    \centering
    \includegraphics[width=1.0\linewidth]{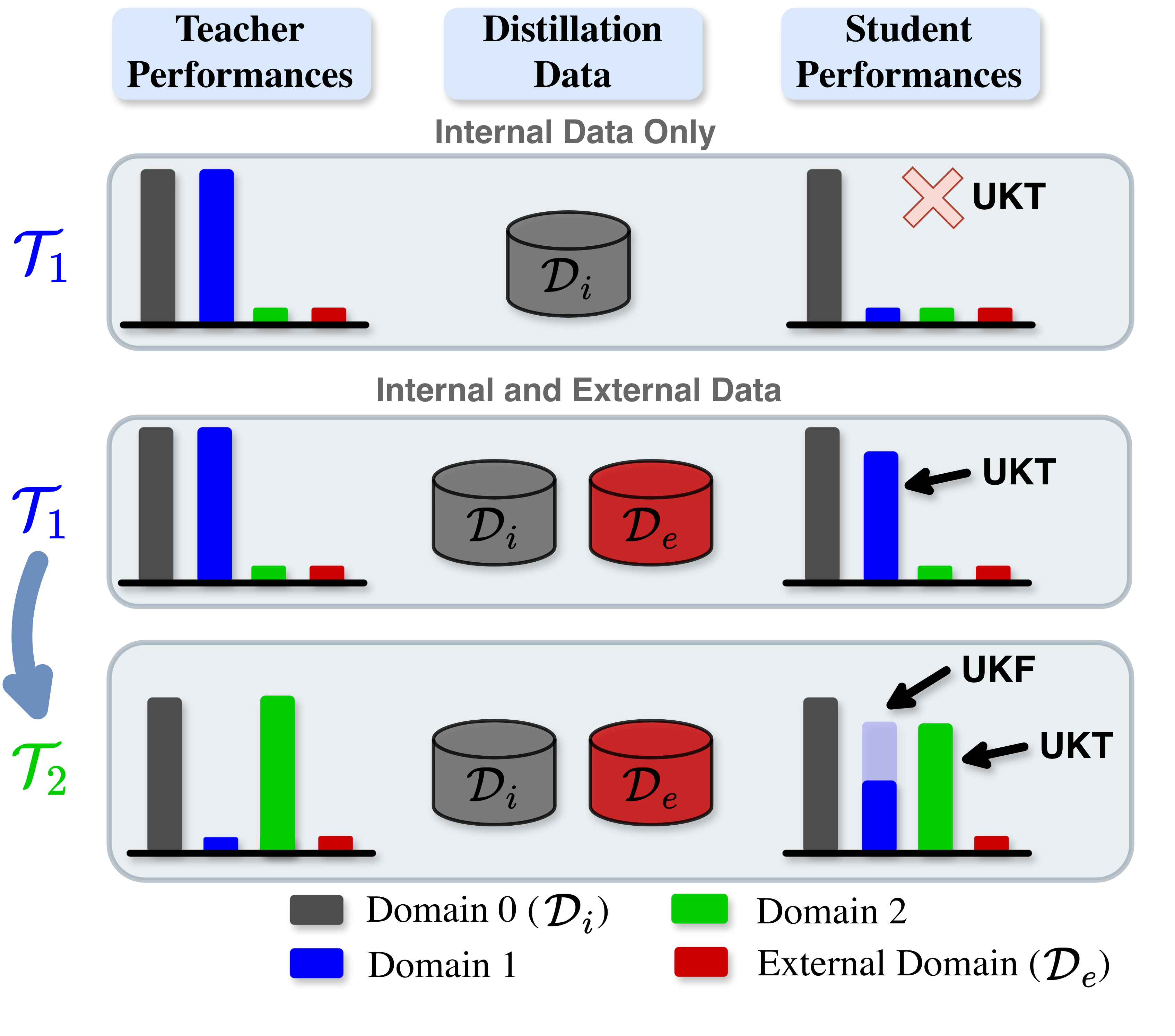}
    \caption{Illustration of Unseen Knowledge Transfer (UKT) and Unseen Knowledge Forgetting (UKF). When distilling only from $\mathcal{T}_1$ on Internal Data $\mathcal{D}_i$, only Domain $0$ knowledge is transferred. When distilling from the same teacher $\mathcal{T}_1$ on $\mathcal{D}_e \cup\mathcal{D}_i$, Domains $0$ and $1$ knowledge is transferred, although the student has never seen Domain $1$. However, when continuing the sequence and distilling from $\mathcal{T}_2$, while the student acquires knowledge from Domain $2$, it forgets part of the knowlege from Domain $1$.}
    \label{fig:ukt}
    % \vspace{-10pt}
\end{figure}
\begin{equation}
\mathcal{D}_t^\mathcal{T} \cap \mathcal{D}_{t^\prime}^\mathcal{T} = \mathcal{D}_i, \  \forall t, t^\prime,
\end{equation}
with $\mathcal{D}_i$ the \textbf{Internal Data (ID)}. The remaining distillation data is considered unknown to all teachers, hence:
\begin{equation}
    \mathcal{D}^{\mathcal{S}} = \mathcal{D}_e \cup \mathcal{D}_i,
\end{equation}
where $\mathcal{D}_e$ denotes the \textbf{External Data (ED)} such that for any teacher of index $t$, $\mathcal{D}_e \cap \mathcal{D}_t = \emptyset$.

\subsection{Unseen Knowledge Transfer and Forgetting}

We define ‘unseen’ as domains not present in the student’s training data but present in the teacher’s knowledge.
In CD, $\mathcal{D}_e$ is unknown to the teachers. Such an ED can either be introduced or can appear unknowingly when generating data, a standard procedure of data-free distillation. While this could appear to be a limitation, we observe that leveraging ED allows the student to acquire knowledge about domains that have never been explicitly seen during training. We refer to this phenomenon as \textbf{Unseen Knowledge Transfer (UKT)}. Intuitively, when integrating ED, generic knowledge is transferred because of the teacher's uncertainty. Conversely, when the teacher is confident, specific knowledge only is transferred. In CD, we propose to take advantage of UKT by purposefully integrating ED during distillation to extract additional knowledge from the teacher.

However, in CD, the student sequentially learns from multiple teachers, each providing distinct unseen domain knowledge. While UKT enables the student to acquire information about domains not directly represented in the teacher data, this transferred knowledge is often fragile. As the student learns from subsequent teachers, they tend to lose information previously transferred from earlier ones. We refer to this phenomenon as \textbf{Unseen Knowledge Forgetting (UKF)}. UKF differs fundamentally from the catastrophic forgetting traditionally studied in Domain Incremental Learning, as the forgotten knowledge does not originate from the student’s own training data but from the teacher’s knowledge. Since the student is never directly exposed to such knowledge, we call it \textit{unseen knowledge}. An intuitive illustration of UKF and UKT is given in Figure~\ref{fig:ukt}.

% Experimentally, we show in Table 1 the impact of distillation data domains with regard to the domains that teachers have been trained on. For the same sequence of teachers $\{\mathcal{T}_{01}, \mathcal{T}_{02}, \mathcal{T}_{03}, \mathcal{T}_{04}\}$, when training on $\mathcal{D}_{0}$ only, performances on domains $\{1, 2, 3\}$ are minimal during all the training process. However, when training on $\mathcal{D}_{04}$, positive performances variations can be observed on $\{1, 2, 3\}$, followed by Unseen Knowledge Forgetting.

\subsection{Self External Data Distillation (SE2D)}
\label{sec:e2d}

\begin{figure}[t]
    \centering
    \includegraphics[width=0.85\linewidth]{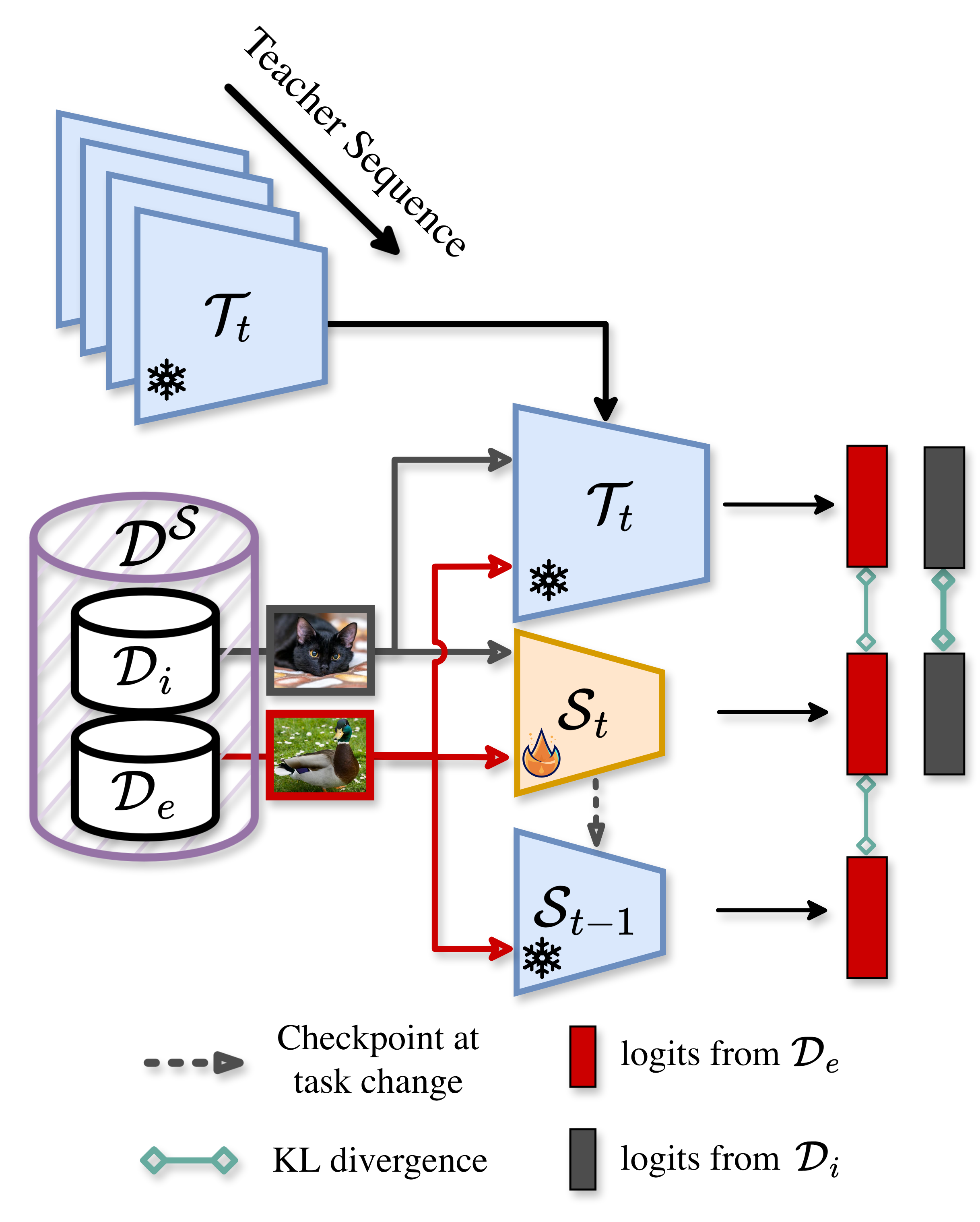}
    \caption{Overview of Self External Data Distillation (SE2D). While the distillation from the teacher sequence is done on the entire distillation data, distillation from the checkpoint of the student at the end of the previous task is done only on $\mathcal{D}_e$.}
    \label{fig:dedo}
\end{figure}

We introduce Self-External Data Distillation (SE2D), a method designed to mitigate UKF in Continual Distillation. In SE2D, the student model is trained not only from the current teacher but also from its own checkpoint saved after the previous task. Such a strategy is quite common in Continual Learning~\cite{li2017learning, rannen2017encoder}; however, we propose to adapt it to the specific problem at hand. Therefore, the distillation process from the checkpoint is performed exclusively on \textit{external data} $\mathcal{D}_e$, which is unknown to all teachers. An overview of our proposed approach is given in Figure~\ref{fig:dedo}.

% \vspace{-10pt}

Prior observations indicate that performance on domains unseen by the student yet known by past teachers depends heavily on these external samples. We restrict self-distillation to external data $\mathcal{D}_e$ to specifically preserve knowledge that is not directly supported by the shared internal domain $\mathcal{D}_i$. Applying self-distillation on $\mathcal{D}_i$ would mainly reinforce already stable knowledge, while our goal is to maintain transferred knowledge from unseen domains, which is primarily captured through $\mathcal{D}_e$.
% By preserving the logits signal of the previous student on $\mathcal{D}_e$, SE2D maintains the semantic continuity of knowledge transferred from earlier teachers, thereby reducing the risk of overwriting it when learning from new ones.

Practically, at each distillation step $t$, the student $\mathcal{S}_{t}$ learns from both the current teacher $\mathcal{T}_{t}$ and the previous student checkpoint $\mathcal{S}_{t-1}$:
\begin{equation}
\mathcal{L}_{\text{SE2D}} = 
\mathcal{L}_{\text{KD}}(\mathcal{S}_{t}, \mathcal{T}_{t}; \mathcal{D}^{\mathcal{S}}) +
\mathcal{L}_{\text{KD}}(\mathcal{S}_{t}, \mathcal{S}_{t-1}; \mathcal{D}_e),
\end{equation}
where $\mathcal{L}_{\text{KD}}$ denotes the temperature-scaled Kullback–Leibler divergence between the softmax distributions of the student logits $z_{\mathcal{S}}(x)$ and teacher logits $z_{\mathcal{T}}(x)$:
\begin{equation}
\label{eq:kl_div}
\mathcal{L}_{\text{KD}}(\mathcal{S}, \mathcal{T}; \mathcal{D}^{\mathcal{S}}) =
T^{2}\,\mathbb{E}_{x \sim \mathcal{D}^{\mathcal{S}}} 
\left[
\text{KL}\left(
\sigma\!\left(\tfrac{z_{\mathcal{T}}(x)}{T}\right)
\;\Big\|\;
\sigma\!\left(\tfrac{z_{\mathcal{S}}(x)}{T}\right)
\right)
\right],
\end{equation}
where $T$ is the distillation temperature and $\sigma(\cdot)$ denotes the softmax function. This simple yet effective mechanism enables the student to accumulate knowledge over time while retaining transferable information from previous distillation stages. In Section~\ref{sec:results}, we present experimental results of SE2D across various benchmarks.

\section{Experimental Setup}
In the following, we describe the experimental setup for Continual Distillation. More details regarding the implementation are given in the appendix.

\subsection{Domain Selection}
\paragraph{Teachers Domains.}
% \vspace{-4pt}
To reproduce a Continual Distillation context, we work on datasets containing data of the same classes but coming from different domains. For example, in Figure~\ref{fig:overview}, each domain contains the class \textit{``cat''}; however, the color of the cat is different depending on the considered domain. Another example would be the condition in which the picture was taken (inside or outside). Such a domain shift is typically considered in DIL~\cite{van2022three, hsu_re-evaluating_2019}.

\paragraph{External Data Selection.}
% \vspace{-4pt}
In CD, the choice of ED is crucial. In realistic scenarios, ED could be introduced accidentally, as knowing FMs exact training domain is unlikely. In this work, we deliberately exploit such external data, specifically selected to fall outside all teacher training domains, to study its influence on knowledge transfer and forgetting. We consider two scenarios: (1) \textbf{related external domains}, where the data share the same semantic classes as the teacher domains (e.g., if teachers are trained on sea animals such as dolphins, sharks, and whales, the external data may include jellyfish); and (2) \textbf{unrelated external domains}, where the data are semantically distinct (e.g., using images of trucks or digits instead of marine animals). In both cases, teachers have minimal performance on ED. Additional discussion is included in appendix.

\subsection{Teacher Selection}
For the experimental setup, we hypothesize that most contemporary FMs share a substantial amount of common knowledge but differ primarily in their performance on specific tasks. Based on this assumption, we train a sequence of teacher models that all share a common domain while each possesses a unique domain not shared by any other teacher. Consequently, all teachers in the sequence provide a shared knowledge base together with a teacher-specific domain that the student must learn and retain through the use of external data. An example is given in Table~\ref{tab:impact_ext} where each teacher is trained on pairs $(0,1)$, $(0,2)$, and so on.

\subsection{Datasets}
For simulating CD, we build upon Domain Incremental Learning datasets. We consider each domain separately and pre-train teachers on subsets of such domains. \textbf{CIFAR20}~\cite{stojanov_incremental_2019}, a variation of CIFAR-100 using the 20 superclasses instead of the 100 fine-grained classes. Since in CIFAR-100, each superclass is composed of 5 sub-classes, using the superclasses allows for defining 5 different domains where each domain is images from different subclasses but identical super-class. CIFAR20 contains $10,000$ train images and $2,000$ test images per domain. All such domains are considered related. Additionally, we experiment with CUB~\cite{wah2011cub} and MNIST as unrelated datasets. \textbf{Digits}, that we define as the combination of various digit datasets, each one representing a different domain. Namely, we mix MNIST~\cite{lecun1998mnist}, MNIST-M~\cite{ganin2016domain}, USPS~\cite{hull2002database} and SVHN~\cite{netzer2011reading}. As a related domain, we consider KMNIST~\cite{clanuwat2018kmnist}, a dataset composed of Japanese Hiragana. We reckon this dataset as having similar difficulty as MNIST, even though of different classes. \textbf{DomainNet}, an adapted version of the DomainNet dataset~\cite{peng2019moment}, containing six visual domains: Real, Clipart, Painting, Infograph, Sketch, and Quickdraw. Each domain shares the same set of 345 classes but differs significantly in style and texture statistics. This dataset contains around 600,000 images, and the domains are unbalanced, increasing the difficulty. All domains are considered related.

\subsection{Considered Methods}
We selected mainstream and recent state-of-the-art baselines for logits-based distillation methods. We considered the following. \textbf{KL-divergence}. This method consists of a standard distillation loss and serves as a baseline~\cite{kl_divergence_book}. \textbf{Logits Standardization (LS)}~\cite{sun2024logit} is a method recently proposed that improves upon logits distillation by standardizing student and teacher logits. \textbf{Medium Difficulty Samples (MDS)}~\cite{chen2025medium} is a data-pruning strategy that considers distilling on samples of medium difficulty only. The original method was proposed in a supervised scenario where the teacher's cross-entropy was considered as the criterion for assessing sample difficulty. In our setup, we adapted this method with the same principle, using teacher entropy as a sample difficulty estimator. \textbf{Decoupled Knowledge Distillation (DKD)}~\cite{zhao2022decoupled}, a standard distillation method that decomposes the conventional distillation objective into two components: a target class term and a non-target class term. While most efficient in supervised scenarios, DKD can easily be used in unsupervised scenarios by considering the teacher's maximum prediction as the target. \textbf{Self-Distillation}. Distillation is regularly used in CL~\cite{michel2024rethinking,douillard2020podnet,rebuffi2017icarl}. A common strategy is to save a checkpoint of the current model at the end of the task and use such a checkpoint for distillation when training on the subsequent task. This methods uses both $\mathcal{D}_i$ and $\mathcal{D}_e$ for distillation.

\section{Experimental Results}
\label{sec:results}
% \vspace{-10pt}
% In the following section, we present the experimental results of training in CD setting. 
% \input{tables/impact_of_no_ext_data}
\input{tables/impact_of_ext_data}

\subsection{External Data Impact on UKT and UKF}
\paragraph{External Data Improves Knowledge Transfer.}
% \vspace{-4pt}

A first observation that can be made in the CD setting is the necessity of training with ED to fully distill knowledge from the teacher. To showcase this effect, we train a sequence of teachers on CIFAR20 on domain pairs $\{0,1\}, \{0,2\}, \{0,3\}$, while distilling to the student model on ID only (domain $0$), using the KL-divergence. The results are displayed in Table~\ref{tab:impact_ext}, where the accuracy of the student after each task is reported. It is important to note that initially, considered teachers achieve above $95\%$ accuracy on their respective domains. Eventually, distilling only on the domain $0$ yields competitive performance on this domain; however, performances on other domains remain extremely limited at any training step. The student performs only on domains that have been encountered during training. In Table~\ref{tab:impact_ext}, we maintain the same teacher sequence but include ED for distillation. Such data are from the domain $4$ of CIFAR20, which is unknown to the teacher. In this case, we can observe that the student maintains performance on domain $0$ while achieving much stronger performances on other domains, despite never being encountered during training. UKT is observed only when distilling with external data.

% \paragraph{External Data Accentuates UKF.}
% % \vspace{-4pt}
% While ED allows for UKT, the majority of the transferred knowledge originates from the last teacher, while previous knowledge is completely forgotten. As shown in Tables~\ref{tab:digit_final} and \ref{tab:domainnet_final_CR}, performance drops on old domains when using ED, except on the last domain seen. For example, for KL-divergence on DomainNet, performances on the \textit{Real} domain drop from $43.43\%$ to $39.32\%$ when leveraging ED. Such results are also observed for other methods and domains, on both Digits and DomainNet. This phenomenon is less pronounced on CIFAR20, occurring only when using MNIST as ED.
\paragraph{External Data Accentuates UKF.} While ED typically facilitates knowledge transfer, it can bias the model toward the most recent teacher at the expense of earlier knowledge. Tables~\ref{tab:cifar_final} and~\ref{tab:digit_final} show that ED does not guarantee uniform gains and may even accentuate forgetting. For instance, on Digits, DKD and LS suffer significant drops on SVHN and MNIST-M; notably, DKD's MNIST-M performance falls from $54.50\%$ to $33.84\%$. However, this effect is not universal, as DKD shows slight gains on DomainNet's Infograph (Table~\ref{tab:domainnet_final_CR}). This suggests that while ED can trigger substantial UKF, the impact depends heavily on the distillation method and domain distribution.
\input{tables/cifar_20_final}
\input{tables/digits_final}
\input{tables/domainnet_final_CR}

\subsection{UKF Mitigation}
% In the following, we discuss the UKF phenomenon as well as the impact of our proposed approach, SE2D.

\paragraph{Traditional Methods and UKF.}
% \vspace{-4pt}

In Tables~\ref{tab:cifar_final}, \ref{tab:digit_final}, and \ref{tab:domainnet_final_CR}, we present the results for all methods in CD setups with related ED. Firstly, it can easily be seen that, as expected, performances on earlier domains tend to be the lowest for all methods, which is a direct demonstration of UKF where the student forgot transferred knowledge on unseen domains. In all scenarios, all traditional distillation methods suffer from UKF. Secondly, Self-Distillation can partially mitigate UKF and surpass distillation approaches due to the regularization effect of self-distillation. However, a clear trade-off emerges between UKT and UKF, not unlike the stability-plasticity trade-off in Continual Learning, where a model has to balance learning and remembering capabilities. Therefore, while self-distillation surpasses most baselines on earlier domains, it struggles to achieve high performances on more recent domains.

\paragraph{Performances of SE2D.}
% \vspace{-4pt}
SE2D allows for better knowledge retention when compared to most baselines, especially on older tasks where it can surpass baselines by more than $9\%$ on domain $1$ of CIFAR20, as reported in Table~\ref{tab:cifar_final}. However, such results hold only for sufficiently related external data where the student can adequately learn from the teacher from one task to another. In Table~\ref{tab:cifar_final}, when using MNIST in combination, the advantages of SE2D become largely limited. This behavior is even more pronounced when training on DomainNet, where SE2D falls behind Self Distillation. We identify the main reason to be the low teacher quality, giving poor supervision on ED for SE2D. Additionally, results in Table~\ref{tab:domainnet_final_CR} show that even when using domains from DomainNet (supposedly related), performance gain is inconsistent for all methods. We believe the vast discrepancy between domains in this dataset hinders UKT, highlighting the importance of the origin of ED for efficient CD. This is discussed in more detail in appendix.

\subsection{Discussion}
\label{sec:discussion}

\paragraph{External Data Ratio Matters.}
Another phenomenon that can be observed in Table~\ref{tab:impact_ext} is that increasing $\frac{|\mathcal{D}_e|}{|\mathcal{D}^\mathcal{S}|}$, the proportion of ED compared to ID, enhances performances on domains unseen by the student. Therefore, a clear trend emerges: the more data unknown to the teacher is used, the stronger the UKT is observed.

\begin{figure}[t]
    \centering
    \includegraphics[width=0.9\linewidth]{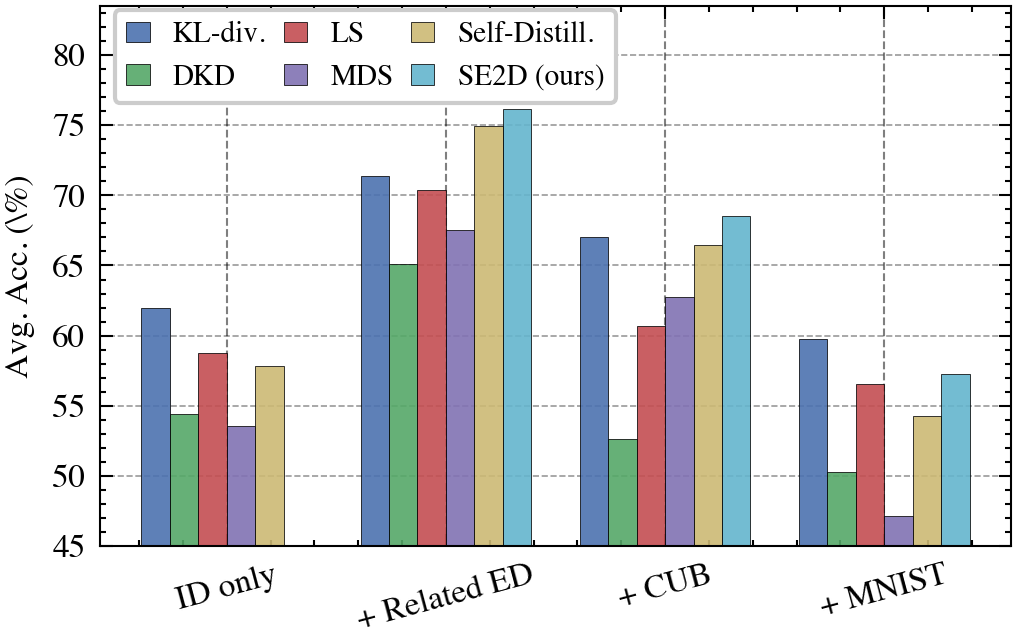}
    \caption{Average Accuracy of the student across all domains known by teachers at the end of training on CIFAR20.}
    \vspace{-5pt}
    \label{fig:barplot}
\end{figure}
% \vspace{-4pt}

\paragraph{External Data Origin Matters.}
% \vspace{-4pt}
Naturally, the origin of the external data has an impact on the intensity of UKT. Intuitively, a large domain difference between external data and teacher domains might make distillation more challenging. To showcase the impact of the domain gap, we experimented with \textbf{related and unrelated external domains} on CIFAR20. Table~\ref{tab:cifar_final} presents the results with various ED scenarios. Notably, it can be observed that leveraging ED leads to consistent improvement in average on all domains when related enough to ID. For example, when using D4 and CUB as ED, the performances of KL-divergence distillation increase from $61.94\%$ to $71.36\%$ and $67.02\%$, respectively. However, when using MNIST as ED, the performances slightly drop to $59.78\%$. Such a trend is observed for all considered methods, as it can be observed in Figure~\ref{fig:barplot}. Semantically, CUB consists of images of birds and is more similar to CIFAR20 than MNIST, even though the number of classes does not align. Interestingly, we observe that leveraging ED is essential to promote UKT. However, as the domain shift between ED and ID increases (e.g., in CUB), performance tends to degrade. When the domain gap becomes too large (as in MNIST), using ED can even result in lower performance than without it.

\paragraph{Limitations of SE2D.}
% \vspace{-4pt}

While SE2D reduces UKF, its impact relies on (1) the domain gap between teacher and external data and (2) teacher performance on domains unseen by the student. SE2D also requires data-origin knowledge; the student must distinguish between the teacher’s known and unknown domains. This is particularly complex when data are generated to imitate training sets, making it non-trivial to identify data outside the teacher's domain.

\section{Conclusions and Future Work}

In this work, we introduced a new paradigm titled Continual Distillation, where a single model learns on a fixed dataset from a sequence of teachers. Such a new setup is relevant in the context of ever-evolving Foundation Models, which are costly to train, expensive to store, demanding to run inference on, and in many cases only accessible via restricted APIs. In such a context, we observe that the domain of origin of distillation data is crucial for controlling which knowledge is indeed transferred to the student. Notably, we unveil two characteristics: Unknown Knowledge Transfer and Unknown Knowledge Forgetting, which represent the ability of the student to modify its knowledge on domains that they have never encountered. Such knowledge control depends only on the teacher and the data used for distillation. The objective then becomes reaching the best UKT-UKF trade-off. In that sense, we proposed Self External Data Distillation (SE2D), which allows us to reduce UKF and maintain strong average performance on all domains, including domains unseen during training. However, we uncover that the domain gap between external data and the teacher domain must be carefully considered in order to foster UKT. Similarly, performant teachers are required for SE2D performance to be ensured.

Eventually, UKT comprises opportunities and risks, as uncontrolled or undesired knowledge could be involuntarily embedded in a student model through distillation depending on the considered data. Such a vulnerability could be easily exploited and introduce unknown bias to model training. Such an aspect of UKT should be explored in future work. Potential future directions include working with larger models, such as language or multimodal models.

\section*{Acknowledgments}
This work was partially financially supported by JST ASPIRE Program, Japan, Grant Number JPMJAP2303. This work was partially supported by the JSPS Postdoctoral Fellowship for Research in Japan (Fellowship ID: P24752).
% \pagebreak

{
    \small
    \bibliographystyle{ieeenat_fullname}
    \bibliography{main}
}

% \pagebreak
% \appendix
\clearpage
\input{appendix}

\end{document}

%% file: preamble.tex
\usepackage[table]{xcolor}

\definecolor{mylightred}{RGB}{253, 231, 231}
\definecolor{mylightyellow}{RGB}{255, 255, 224}
\definecolor{mylightgreen}{RGB}{229, 245, 229}
\definecolor{mylightblue}{RGB}{227, 235, 255}

%% file: tables/impact_of_ext_data.tex
\begin{table}[ht]
    \small
    \centering
    \caption{Accuracy (\%) of the student model on test sets, domain-wise, on CIFAR20, for various domain overlaps, after distilling for $1$ epoch. The larger the ratio $|\mathcal{D}_e|/|\mathcal{D}^\mathcal{S}|$, the more the student performs on domains unseen during training (underlined values).}
    \vspace{-5pt}
        \centering
        \resizebox{0.47\textwidth}{!}{
        \begin{tabular}{l||c|c|c|c||c} 
        \toprule
        Domain   & \textbf{0} & \textbf{1} & \textbf{2} & \textbf{3} & $|\mathcal{D}_e|/|\mathcal{D}^\mathcal{S}|$ \\
        \midrule \midrule
        After $\mathbb{D}(\mathcal{T}_{01}, \mathcal{D}^{\mathcal{S}}_0)$ & \textbf{93.25} & \underline{37.80} & 45.30 & 36.90 & 0\%\\
        After $\mathbb{D}(\mathcal{T}_{02}, \mathcal{D}^{\mathcal{S}}_0)$ & \textbf{93.95} & 31.60 & \underline{42.20} & 35.25& 0\% \\
        After $\mathbb{D}(\mathcal{T}_{03}, \mathcal{D}^{\mathcal{S}}_0)$ & \textbf{92.10} & 32.00 & 39.05 & \underline{33.00} & 0\%\\
        \midrule \midrule
        After $\mathbb{D}$($\mathcal{T}_{012}$, $\mathcal{D}^{\mathcal{S}}_{014}$) & \textbf{92.70} & \textbf{93.15} & \underline{68.35} & 53.80 & 33\%\\
        % \midrule
        After $\mathbb{D}$($T_{013}$, $\mathcal{D}^{\mathcal{S}}_{014}$) & \textbf{92.45} & \textbf{92.95} & 48.95 & \underline{72.55} & 33\%\\
        \midrule \midrule
        After $\mathbb{D}$($\mathcal{T}_{01}$, $\mathcal{D}^{\mathcal{S}}_{04}$) &  \textbf{96.35} & \underline{77.15} & 48.95 & 48.45 & 50\% \\
        % \midrule
        After $\mathbb{D}$($\mathcal{T}_{02}$, $\mathcal{D}^{\mathcal{S}}_{04}$) &  \textbf{96.35} & 43.30 & \underline{80.10} & 46.55 & 50\% \\
        % \midrule
        After $\mathbb{D}$($\mathcal{T}_{03}$, $\mathcal{D}^{\mathcal{S}}_{04}$) &  \textbf{95.70} & 49.40 & 57.60 & \underline{77.80} & 50\% \\
        \midrule \midrule
        After $\mathbb{D}$($\mathcal{T}_{01}$, $\mathcal{D}^{\mathcal{S}}_{034}$) & \textbf{94.60} & \underline{85.20} & 51.85 & 57.15  & 66\% \\
        % \midrule
        After $\mathbb{D}$($\mathcal{T}_{02}$, $\mathcal{D}^{\mathcal{S}}_{034}$) & \textbf{94.60} & 44.00 & \underline{83.55} & 51.55 & 66\% \\
        \bottomrule
        \end{tabular}}
    \vspace{-10pt}
    \label{tab:impact_ext}
\end{table}

%% file: tables/cifar_20_final.tex
\begin{table*}[t]
\centering
\renewcommand{\arraystretch}{1}
\setlength{\tabcolsep}{5pt}

% --- Define colors ---
\definecolor{OutDomCol}{RGB}{255,178,178}   % light red
\definecolor{InDomCol}{RGB}{211,211,211}    % light gray
\definecolor{DomainCol}{RGB}{255,255,255}   % light gray
\definecolor{AvgCol}{RGB}{250,235,220}      % light orange
\definecolor{OtherDom}{RGB}{230,240,250}   % light blue

% --- Table ---
\caption{Performances (\%, higher is better) of the student at the end of training on CIFAR20 for 4 scenarios. Internal Data Only (D0), Related External Data (D4), CUB as ED, and MNIST as ED. The number of runs is set to $3$. The gain columns shows the gain over using Internal Data Only. \colorbox{InDomCol}{Grey}: Internal Data; \colorbox{OtherDom}{Blue}: Domain known by the teacher; \colorbox{OutDomCol}{Red}: External Data (ED); White: Ignored.}
\vspace{-4pt}
\resizebox{0.76\textwidth}{!}{%
\begin{tabular}{
    >{\columncolor{DomainCol}\color{black}}l
    >{\columncolor{InDomCol}}c
    *{3}{>{\columncolor{OtherDom}\color{black}}c}
    >{\columncolor{DomainCol}}c
    >{\columncolor{AvgCol}}c
    >{\columncolor{DomainCol}\color{black}}c
}
\toprule
  \multicolumn{8}{c}{\textbf{CIFAR20 - Internal Data Only}} \\
  \hline
 \multicolumn{1}{c}{\textbf{Method}} & D0 & D1 & D2 & D3 & D4 \xmark & Avg. (0-3) & Gain ($\uparrow$) \\
\hline
$\mathcal{T}_{best}$ (upper bound) & 97.75 & 95.80 & 96.70 & 95.75 & - & 96.5 & 0.00 \\
KL-divergence & \textbf{98.10 \std{0.05}} & \textbf{41.40 \std{0.84}} & \textbf{53.38 \std{0.35}} & \textbf{54.87 \std{3.70}} & - & \textbf{61.94 \std{1.24}} & 0.00 \\
DKD [CVPR'22] & 95.70 \std{1.55} & 35.53 \std{1.94} & 45.42 \std{2.05} & 41.02 \std{2.55} & - & 54.42 \std{2.02} & 0.00 \\
LS [CVPR'24] & 96.50 \std{1.52} & \underline{39.13 \std{2.90}} & 49.77 \std{4.40} & \underline{49.58 \std{5.78}} & - & \underline{58.75 \std{3.65}} & 0.00 \\
MDS [ICLR'25] & 94.20 \std{0.73} & 34.30 \std{2.80} & 44.70 \std{1.40} & 41.00 \std{1.90} & - & 53.55 \std{1.72} & 0.00 \\
Self-Distillation & \underline{97.45 \std{0.43}} & 37.58 \std{2.06} & \underline{50.42 \std{1.41}} & 45.83 \std{2.15} & \makebox[\widthof{73.65 \std{1.67}}][c]{-} & 57.82 \std{1.51} & 0.00 \\
\end{tabular}%
}

\resizebox{0.76\textwidth}{!}{%
\begin{tabular}{
    >{\columncolor{DomainCol}\color{black}}l
    >{\columncolor{InDomCol}}c
    *{3}{>{\columncolor{OtherDom}\color{black}}c}
    >{\columncolor{OutDomCol}}c
    >{\columncolor{AvgCol}}c
    >{\columncolor{DomainCol}\color{black}}c
}
\toprule
\multicolumn{8}{c}{\textbf{CIFAR20 + Related External Data}} \\
\hline
 \multicolumn{1}{c}{\textbf{Method}}  & D0 & D1 & D2 & D3 & D4  & Avg. (0-3) & Gain ($\uparrow$) \\
\hline
$\mathcal{T}_{best}$ (upper bound) & 97.75 & 95.80 & 96.70 & 95.75 & - & 96.5 & 0.00 \\
KL-divergence & 97.05 \std{0.09} & 48.55 \std{1.15} & 55.08 \std{0.70} & \textbf{84.77 \std{0.87}} & - & 71.36 \std{0.70} & 9.42 \\
DKD [CVPR'22] & 96.05 \std{0.50} & 44.13 \std{0.98} & 51.67 \std{0.92} & 68.55 \std{2.60} & - & 65.10 \std{1.25} & 10.68 \\
LS [CVPR'24] & 96.85 \std{0.15} & 47.25 \std{0.69} & 54.25 \std{0.46} & \underline{83.20 \std{1.87}} & - & 70.39 \std{0.79} & 11.64 \\
MDS [ICLR'25] & 96.55 \std{0.07} & 45.26 \std{2.10} & 54.90 \std{0.71} & 73.51 \std{0.56} & - & 67.56 \std{0.86} & 14.01 \\
Self-Distillation & \textbf{97.71 \std{0.18}} & \underline{61.23 \std{0.83}} & \textbf{64.21 \std{0.51}} & 76.58 \std{0.94} & - & \underline{74.93 \std{0.61}} & 17.11 \\
\hline
SE2D (ours) & \underline{97.46 \std{0.19}} & \textbf{70.71 \std{1.05}} & \underline{62.85 \std{0.50}} & 73.65 \std{1.67} & \makebox[\widthof{73.65 \std{1.67}}][c]{-} & \textbf{76.17 \std{0.85}} & n/a \\
\end{tabular}%
}

\resizebox{0.76\textwidth}{!}{%
\begin{tabular}{
    >{\columncolor{DomainCol}\color{black}}l
    >{\columncolor{InDomCol}}c
    *{3}{>{\columncolor{OtherDom}\color{black}}c}
    >{\columncolor{OutDomCol}}c
    >{\columncolor{AvgCol}}c
    >{\columncolor{DomainCol}\color{black}}c
}
\toprule
 \multicolumn{8}{c}{\textbf{CIFAR20 + CUB}} \\
 \hline
 \multicolumn{1}{c}{\textbf{Method}} & D0 & D1 & D2 & D3 & CUB & Avg. (0-3) & Gain ($\uparrow$) \\
\hline
$\mathcal{T}_{best}$ (upper bound) & 97.75 & 95.80 & 96.70 & 95.75 & - & 96.5 & 0.00 \\
KL-divergence  & 97.24 \std{0.37} & 43.89 \std{1.02} & 55.13 \std{0.76} & \textbf{71.80 \std{1.73}} & - & \underline{67.02 \std{0.97}} & 5.08 \\
DKD [CVPR'22]  & 93.39 \std{0.88} & 33.46 \std{2.51} & 43.10 \std{2.93} & 40.70 \std{2.20} & - & 52.66 \std{2.13} & -1.76 \\
LS [CVPR'24] & 94.79 \std{3.17} & 38.53 \std{4.87} & 50.21 \std{5.74} & 59.32 \std{11.60} & - & 60.71 \std{6.34} & 1.96 \\
MDS [ICLR'25]& 97.15  \std{0.50} &  41.12  \std{0.96}  & 52.38  \std{1.76} &  60.38  \std{2.48} & - &   62.76  \std{0.89} & 9.21 \\
Self-Distillation  & \underline{97.47 \std{0.12}} & \underline{47.97 \std{1.07}} & \textbf{58.40 \std{1.34}} & 61.97 \std{1.82} & - & 66.45 \std{1.09} & 8.63 \\
\hline
SE2D (ours) & \textbf{97.74 \std{0.10}} & \textbf{53.93 \std{0.43}} & \underline{58.02 \std{0.45}} & \underline{64.54 \std{1.81}} & \makebox[\widthof{73.65 \std{1.67}}][c]{-} & \textbf{68.56 \std{0.70}} & n/a \\
\midrule
 \multicolumn{8}{c}{\textbf{CIFAR20 + MNIST}} \\
 \hline
 \multicolumn{1}{c}{\textbf{Method}} & D0 (ID) & D1 & D2 & D3  & MNIST & Avg. (0-3) & Gain ($\uparrow$) \\
\hline
$\mathcal{T}_{best}$ (upper bound) & 97.75 & 95.80 & 96.70 & 95.75 & - & 96.5 & 0.00 \\
KL-divergence & \textbf{94.45 \std{4.20}} & \underline{38.24 \std{6.06}} & \textbf{48.44 \std{8.56}} & \textbf{57.97 \std{15.71} }& - & \textbf{59.78 \std{8.63}} & -2.16 \\
DKD [CVPR'22]   & 91.00 \std{3.35} & 30.71 \std{3.60} & 41.36 \std{3.61} & 37.96 \std{4.56} & - & 50.26 \std{3.78} & -4.16 \\
LS [CVPR'24]  & 92.04 \std{4.43} & 35.93 \std{6.37} & 45.79 \std{7.50} & \underline{52.42 \std{13.24}} & - & 56.54 \std{7.88} & -2.21 \\
MDS [ICLR'25] & 86.83 \std{1.04} & 28.53 \std{2.20} &   37.92 \std{2.41} &   37.38 \std{3.24} & - & 47.17 \std{1.57} & -6.38 \\
Self-Distillation & 91.90 \std{4.94} & 36.35 \std{9.46} & 45.23 \std{9.18} & 43.58 \std{12.43} & - & 54.26 \std{9.00} & -3.56 \\
\hline
SE2D (ours)  & \underline{92.64 \std{4.66}} & \textbf{39.94 \std{9.73}} & \underline{47.55 \std{9.55}} & 48.88 \std{13.23} & \makebox[\widthof{73.65 \std{1.67}}][c]{-} & \underline{57.25 \std{9.29}} & n/a \\
\bottomrule
\end{tabular}%
}
\label{tab:cifar_final}
\end{table*}

%% file: tables/digits_final.tex
\begin{table*}[t]
\centering
\renewcommand{\arraystretch}{1}
\setlength{\tabcolsep}{5pt}

% --- Define colors ---
\definecolor{OutDomCol}{RGB}{255,178,178}   % light red
\definecolor{InDomCol}{RGB}{211,211,211}    % light gray
\definecolor{DomainCol}{RGB}{255,255,255}   % light gray
\definecolor{AvgCol}{RGB}{250,235,220}      % light orange
\definecolor{OtherDom}{RGB}{230,240,250}   % light blue

% --- Table ---
\caption{Performances (\%, higher is better) of the student at the end of training on Digits for 2 scenarios. Internal Data Only (MNIST) and Related External Data (KMNIST). The number of runs is set to $3$. Average and standard deviations are reported.}
\vspace{-4pt}
\resizebox{0.76\textwidth}{!}{%
\begin{tabular}{
    >{\columncolor{DomainCol}\color{black}}l
    >{\columncolor{InDomCol}}c
    *{3}{>{\columncolor{OtherDom}\color{black}}c} 
    >{\columncolor{DomainCol}}c
    >{\columncolor{AvgCol}}c
    >{\columncolor{DomainCol}\color{black}}c
}
\toprule
 \multicolumn{8}{c}{\textbf{Digits - Internal Data Only}} \\
\hline
 \multicolumn{1}{c}{\textbf{Method}} & MNIST & SVHN & MNIST-M & USPS & KMNIST \xmark  & Avg. & Gain ($\uparrow$) \\
\hline
$\mathcal{T}_{best}$ (upper bound) & 99.2 & 97.84 & 99.25 & 98.8 & - & 98.77 & 0.00 \\
KL-divergence & \underline{99.17 \std{0.04}} & \underline{35.80 \std{1.49}} & 62.80 \std{2.69} &\underline{95.81 \std{0.46}} & - & 73.40 \std{1.17} & 0.00 \\
DKD [CVPR'22] & 98.70 \std{0.11} & 33.35 \std{1.28} & 54.50 \std{2.45} & 95.07 \std{0.43} & - & 70.40 \std{1.07} & 0.00 \\
LS [CVPR'24] & 99.13 \std{0.09} & \textbf{37.60 \std{1.40}} & \underline{64.10 \std{1.73}} & \textbf{96.00 \std{0.25}} & - & \textbf{74.21 \std{0.87}} & 0.00 \\
MDS [ICLR'25] & 98.15 \std{1.13} &  34.51 \std{4.80} &  54.97 \std{9.49} &  93.21 \std{3.01} & - & 70.21 \std{3.80} & 0.00 \\
Self-Distillation & \textbf{99.23 \std{0.10}} & 35.08 \std{1.03} & \textbf{65.87 \std{1.38}} & 95.28 \std{0.53} & \makebox[\widthof{73.65 \std{1.67}}][c]{-} & \underline{73.87 \std{0.76}} & 0.00 \\
\end{tabular}}
\resizebox{0.76\textwidth}{!}{%
\begin{tabular}{
    >{\columncolor{DomainCol}\color{black}}l
    >{\columncolor{InDomCol}}c
    *{3}{>{\columncolor{OtherDom}\color{black}}c}
    >{\columncolor{OutDomCol}}c
    >{\columncolor{AvgCol}}c
    >{\columncolor{DomainCol}\color{black}}c
}
\toprule
 \multicolumn{8}{c}{\textbf{Digits + Related External Data}} \\
 \hline
 \multicolumn{1}{c}{\textbf{Method}}  & MNIST & SVHN & MNIST-M & USPS & KMNIST & Avg. & Gain ($\uparrow$) \\
\hline
% $\mathcal{T}_{best}$ (upper bound) & 99.2 & 97.84 & 99.25 & 98.8 & - & 98.77 & 0.00 \\
KL-divergence & 99.13 \std{0.05} & 31.53 \std{1.55} & 59.84 \std{2.57} & \textbf{96.51 \std{0.10}} & - & 71.75 \std{1.07} & -1.65 \\
DKD [CVPR'22] & 98.35 \std{0.32} & 25.21 \std{1.62} & 33.84 \std{4.28} & 92.87 \std{1.63} & - & 62.57 \std{1.96} & -7.83 \\
LS [CVPR'24] & 99.13 \std{0.08} & 32.28 \std{0.37} & 61.47 \std{2.15} & 96.33 \std{0.33} & - & 72.30 \std{0.73} & -1.91 \\
MDS [ICLR'25] & 99.12 \std{0.04} & 33.03 \std{1.79} & 60.74 \std{0.97} & 96.13 \std{0.05} & - & 62.50 \std{0.90} & -7.71 \\
Self-Distillation & \textbf{99.38 \std{0.01}} & \underline{55.86 \std{1.60}} & \textbf{90.76 \std{0.35}} & 96.33 \std{0.15} & - & \underline{85.58 \std{0.53}} & 11.71 \\
\hline
SE2D (ours) & \underline{99.33} \std{0.04} & \textbf{61.84 \std{2.05}} & \underline{90.44 \std{0.18}} & \underline{96.33 \std{0.10}} & \makebox[\widthof{73.65 \std{1.67}}][c]{-} & \textbf{87.00 \std{0.60}} & n/a \\
\bottomrule
\end{tabular}%
}
\label{tab:digit_final}
\end{table*}

%% file: tables/domainnet_final_CR.tex
\begin{table*}[t]
\centering
\renewcommand{\arraystretch}{1}
\setlength{\tabcolsep}{5pt}

% --- Define colors ---
\definecolor{OutDomCol}{RGB}{255,178,178}   % light red
\definecolor{InDomCol}{RGB}{211,211,211}    % light gray
\definecolor{DomainCol}{RGB}{255,255,255}   % white
\definecolor{AvgCol}{RGB}{250,235,220}      % light orange
\definecolor{OtherDom}{RGB}{230,240,250}    % light blue

% --- Table ---
\caption{Performances (\%, higher is better) of the student at the end of training on DomainNet for 2 scenarios. Internal Data Only (Clipart) and Related External Data (Sketch). The number of runs is set to $3$. Average and standard deviations are reported.}
\vspace{-4pt}
\resizebox{0.85\textwidth}{!}{%
\begin{tabular}{
    >{\columncolor{DomainCol}\color{black}}l
    >{\columncolor{InDomCol}}c
    *{4}{>{\columncolor{OtherDom}}c}   
    >{\columncolor{DomainCol}}c
    >{\columncolor{AvgCol}}c
    >{\columncolor{DomainCol}\color{black}}c
}
\toprule
 & \multicolumn{7}{c}{\textbf{DomainNet - Internal Data Only}} \\
 \hline
 \multicolumn{1}{c}{\textbf{Method}} & Clipart & Infograph & Painting & Quickdraw & Real & Sketch \xmark  & Avg. & Gain ($\uparrow$) \\
\hline
$\mathcal{T}_{best}$ (upper bound) & 74.35 & 35.83 & 66.66 & 66.19 & 78.90 & - & 64.39 & 0.00 \\
KL-divergence & 77.08 \std{0.45} & \underline{18.16 \std{0.24}} & \underline{44.17 \std{0.50}} & 17.19 \std{0.73} & \textbf{67.29 \std{0.24}} & \makebox[\widthof{73.65 \std{1.67}}][c]{-} & \underline{44.78 \std{0.29}} & 0.00 \\
DKD [CVPR'22] & \underline{77.32 \std{0.22}} & 18.04 \std{0.09} & 42.76 \std{0.40} & 17.43 \std{0.31} & 63.99 \std{0.14} & \makebox[\widthof{73.65 \std{1.67}}][c]{-} & 43.91 \std{0.17} & 0.00 \\
LS [CVPR'24] & 77.21 \std{0.37} & 17.21 \std{0.91} & 40.95 \std{0.14} & 16.42 \std{1.01} & 60.39 \std{0.20} & \makebox[\widthof{73.65 \std{1.67}}][c]{-} & 42.43 \std{0.44} & 0.00 \\
MDS [ICLR'25] & 76.92 \std{0.42} & 18.08 \std{0.23} & 42.14 \std{0.38} & \underline{17.44 \std{1.03}} & 63.29 \std{0.20} & \makebox[\widthof{73.65 \std{1.67}}][c]{-} & 43.57 \std{0.34} & 0.00 \\
Self-Distillation & \textbf{80.57 \std{0.08}} & \textbf{20.85 \std{0.33}} & \textbf{46.23 \std{0.11}} & \textbf{23.53 \std{0.25}} & \underline{65.38 \std{0.04}} & \makebox[\widthof{73.65 \std{1.67}}][c]{-} & \textbf{47.31 \std{0.11}} & 0.00 \\
\end{tabular}
}
\vspace{5pt}
\resizebox{0.85\textwidth}{!}{%
\begin{tabular}{
    >{\columncolor{DomainCol}\color{black}}l
    >{\columncolor{InDomCol}}c
    *{4}{>{\columncolor{OtherDom}}c}   
    >{\columncolor{OutDomCol}}c
    >{\columncolor{AvgCol}}c
    >{\columncolor{DomainCol}\color{black}}c
}
\toprule
 & \multicolumn{7}{c}{\textbf{DomainNet + Related External Data}} \\
\hline
\multicolumn{1}{c}{\textbf{Method}} & Clipart & Infograph & Painting & Quickdraw & Real & Sketch & Avg. & Gain ($\uparrow$) \\
\hline
% $\mathcal{T}_{best}$ (upper bound) & 74.35 & 35.83 & 66.66 & 66.19 & 78.90 & - & 64.39 & 0.00 \\
KL-divergence & 76.00 \std{0.18} & 18.89 \std{0.09} & 44.77 \std{0.79} & 15.53 \std{0.15} & \textbf{70.65 \std{0.46}} & \makebox[\widthof{73.65 \std{1.67}}][c]{-} & 45.17 \std{0.31} & 0.39 \\
DKD [CVPR'22] & 76.53 \std{0.14} & 18.70 \std{0.44} & 44.24 \std{0.32} & 16.24 \std{0.40} & 68.29 \std{0.34} & \makebox[\widthof{73.65 \std{1.67}}][c]{-} & 44.80 \std{0.12} & 0.89 \\
LS [CVPR'24] & 75.28 \std{0.39} & 16.88 \std{0.71} & 40.78 \std{0.49} & 15.47 \std{0.43} & 62.76 \std{0.64} & \makebox[\widthof{73.65 \std{1.67}}][c]{-} & 42.24 \std{0.31} & -0.20 \\
MDS [ICLR'25] & 75.32 \std{0.39} & 17.53 \std{0.07} & 42.28 \std{0.11} & 16.12 \std{0.61} & 67.29 \std{0.11} & \makebox[\widthof{73.65 \std{1.67}}][c]{-} & 43.71 \std{0.19} & 0.14 \\
Self-Distillation & \textbf{80.10 \std{0.18}} & \underline{21.53 \std{0.09}} & \textbf{48.15 \std{0.23}} & \textbf{25.28 \std{0.19}} & \underline{68.75 \std{0.01}} & \makebox[\widthof{73.65 \std{1.67}}][c]{-} & \textbf{48.76 \std{0.07}} & 1.45 \\
\hline
SE2D (ours) & \underline{78.05 \std{0.11}} & \textbf{21.98 \std{0.29}} & \underline{47.76 \std{0.44}} & \underline{23.81 \std{0.34}} & 68.43 \std{0.14} & \makebox[\widthof{73.65 \std{1.67}}][c]{-} & \underline{48.01 \std{0.20}} & n/a \\
\bottomrule
\end{tabular}
}
\vspace{-4pt}
\label{tab:domainnet_final_CR}
\end{table*}

%% file: appendix.tex
\appendix

\counterwithin{figure}{section}
\counterwithin{table}{section}
\counterwithin{equation}{section}

\section{Experimental Setup}
\subsection{Implementation Details}

For training, we start from pre-trained weights and use the Adam optimizer with a learning rate of $0.0001$ for 3 epochs. As students start from pre-trained weights, we observe negligible improvement when using more epochs. Images are resized to $224\times224$ to fit the size used during pre-training. We use random horizontal flips as augmentations and use data normalization. We use a batch size of 64. Regarding the distillation, we use the KL-divergence with a temperature of $10$. Experiments are conducted with ViT architecture, namely the ViT-B/16. We use ViT-base as teachers for CIFAR20 and DomainNet. For Digits, we use a ViT-tiny as a teacher. Every teacher is initialised from pre-trained weights and trained for $50$ epochs with Adam optimizer and a learning rate of $0.0001$. We use the same architectures for students in the main draft and additionally experiment with ViT-tiny. For all results, 3 seeds are used.

For DomainNet, since domains are unbalanced, we oversample or undersample them so that each task has the same number of steps. The base task length is determined by the Internal Data (ID) size, and the External Data (ED) is sampled accordingly. For example, if ID is larger than ED, we oversample from ED.

\subsection{Metric}
For all experiments, we report the domain-wise Final Accuracy ($\%$) of the student at the end of the training sequence.

\section{Additional Discussions}
\paragraph{Feasibility of obtaining unknown ED.} In CD, a question naturally arises as to whether obtaining ED unknown to FMs is feasible, since such models are trained on vast and diverse data. We argue that obtaining unknown data is feasible in two highly plausible scenarios: \textbf{Temporal Gap:} Any data uploaded to public repositories \textit{after} the FM's training cutoff is guaranteed to be unknown. \textbf{Private/Synthetic Data:} In industrial settings, private proprietary datasets or synthetically generated data serve as excellent ED.

\begin{figure}[t]
    \centering
    \includegraphics[width=1.00\linewidth]{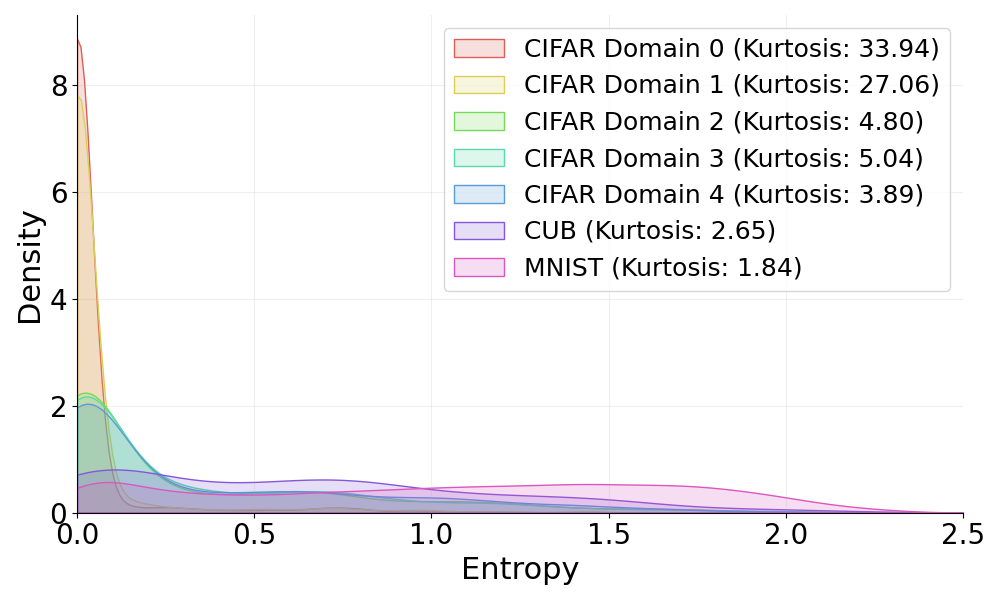}
    \caption{Entropy distribution of the teacher (ViT-Base) trained on domains 0 and 1 of \textbf{CIFAR20} for various datasets and domains.}
    \label{fig:entropy_cifar}
\end{figure}

\begin{figure}[t]
    \centering
    \includegraphics[width=1.00\linewidth]{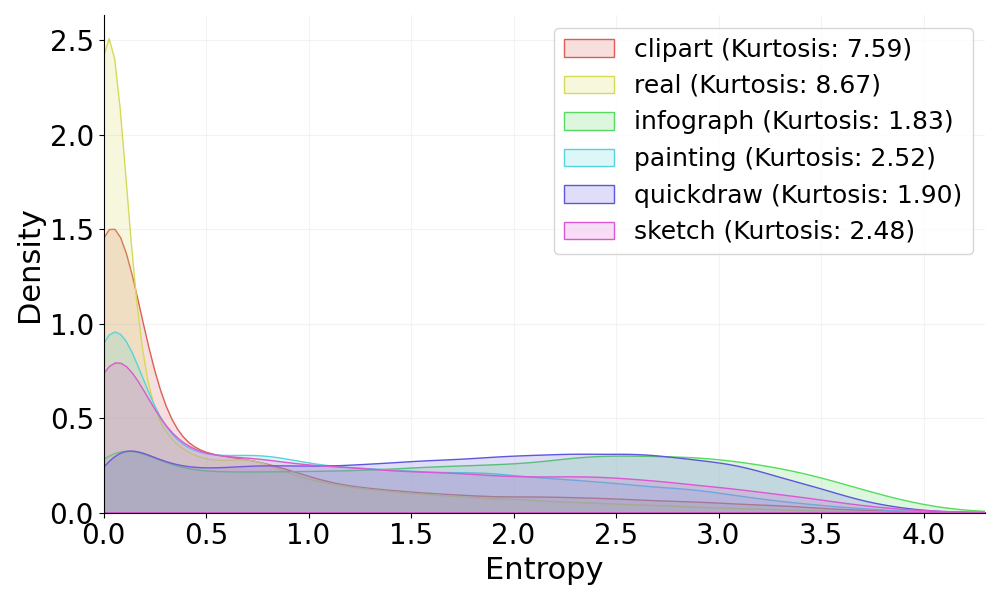}
    \caption{Entropy distribution of the teacher (ViT-Base) trained on domains Real and Clipart of Domainnet for various \textbf{DomainNet} domains.}
    \label{fig:entropy_domainnet}
\end{figure}

\paragraph{Assessing ED quality.}
As presented throughout this paper, ED selection is key in Continual Distillation. While quantifying semantic similarity is hard without teacher data, we propose using the teacher's own predictive uncertainty, by measuring the entropy, as a proxy. In Figure~\ref{fig:entropy_cifar}, we show the entropy distribution of a teacher trained on two domains of CIFAR20, for various domains of CIFAR20 as well as CUB and MNIST. We observe that \textbf{the more the external data is "unrelated", the more "flat" the entropy distribution becomes}. To select adequate ED without knowing the training distribution, a user can 1) filter out samples based on an entropy threshold, 2) use the 4th-order moment (kurtosis) of the entropy distribution to quantify "flatness". As shown in Fig.~\ref{fig:entropy_cifar}, lower flatness correlates with higher UKT potential. Additionally, in Figure~\ref{fig:entropy_domainnet}, the entropy distribution varies widely across domains, partially explaining the mitigated results observed on this dataset.

\section{Additional Experiments}
\subsection{Additional Metrics}
\paragraph{Forgetting} 
For each method, we measure forgetting as the drop in performance on the learned domains after training from new teachers. Formally, for a student trained to imitate a teacher of index $t$, the forgetting for domain $d$ is computed as:

\[
F_d = \max_{i < t} A_{d}^{(i)} - A_{d}^{(t)}
\]

where $A_{d}^{(i)}$ is the accuracy on domain $d$ after distillation from teacher $t$. The overall forgetting is averaged across all domains:

\[
F = \frac{1}{D} \sum_{d=1}^{D} F_d
\]

This metric captures the extent to which a method forgets previously learned knowledge when adapting to new tasks. The results are presented in Tables~\ref{tab:fgt_c20}, \ref{tab:fgt_digit} and \ref{tab:fgt_domainnet_cr}. It can be observed that our method leads to competitive forgetting in all scenarios.

\input{tables/cifar_20_all_forgetting}

\begin{algorithm}[h]
\caption{Continual Distillation with KL divergence and SGD.}
\label{algo:cd}
\begin{algorithmic}[1]
\REQUIRE Sequence of teachers $\{\mathcal{T}_1, \mathcal{T}_2, \dots, \mathcal{T}_T\}$, student model $\mathcal{S}_\theta$, distillation dataset $\mathcal{D}^\mathcal{S}$
\FOR{$t = 1$ to $N$}
    \FOR{$x \in \mathcal{D}^\mathcal{S}$}
        \STATE Obtain teacher predictions $p_t(x) = \mathcal{T}_t(x)$
        \STATE Student predictions $q_\theta(x) = \mathcal{S}_\theta(x)$
        \STATE Distillation loss: $\mathcal{L}_t = \mathrm{KL}\big(p_t(x) \,\|\, q_\theta(x)\big)$
        \STATE Update student parameters $\theta \leftarrow \theta - \eta \nabla_\theta \mathcal{L}_t$
    \ENDFOR
\ENDFOR
\RETURN Trained student model $\mathcal{S}_\theta$
\end{algorithmic}
\end{algorithm}

\begin{algorithm*}[h]
\caption{Overview of SE2D training algorithm with Continual Distillation.}
\label{alg:ours}
\begin{algorithmic}[1]
\REQUIRE Sequence of teachers $\{\mathcal{T}_1, \mathcal{T}_2, \dots, \mathcal{T}_T\}$, student model $\mathcal{S}_\theta$, distillation dataset $\mathcal{D}^\mathcal{S} = \mathcal{D}_i \cup \mathcal{D}_e$
\FOR{$t = 1$ to $N$}
    \IF{$t = 1$}
        \FOR{$x \in \mathcal{D}^\mathcal{S}$}
            \STATE Obtain teacher predictions $p_t(x) = \mathcal{T}_t(x)$
            \STATE Compute student predictions $q_\theta(x) = \mathcal{S}_\theta(x)$
            \STATE Compute distillation loss: $\mathcal{L}_t = \mathrm{KL}\big(p_t(x) \,\|\, q_\theta(x)\big)$
            \STATE Update student parameters $\theta \leftarrow \theta - \eta \nabla_\theta \mathcal{L}_t$
        \ENDFOR
    \ELSE
        \STATE Load previous student checkpoint $\mathcal{S}_{\theta}^{t-1}$
        \FOR{$(x_e, x_i) \in (\mathcal{D}_e, \mathcal{D}_i)$}
            \STATE Compute student predictions on internal data: $q_\theta(x_i) = \mathcal{S}_\theta(x_i)$
            \STATE Compute student predictions on external data: $q_\theta(x_e) = \mathcal{S}_\theta(x_e)$
            \STATE Obtain previous student predictions on external data: $p_{t-1}(x_e) = \mathcal{S}_{\theta}^{t-1}(x_e)$
            \STATE Obtain current teacher predictions on all data: $p_t^{\text{teacher}}((x_e, x_i)) = \mathcal{T}_t((x_e, x_i))$
            \STATE Compute distillation loss on external data from previous student: $\mathcal{L}_{\text{student}} = \mathrm{KL}\big(p_{t-1}(x_e) \,\|\, q_\theta(x_e)\big)$
            \STATE Compute distillation loss on all data from teacher: $\mathcal{L}_{\text{teacher}} = \mathrm{KL}\big(p_t^{\text{teacher}}((x_e, x_i)) \,\|\, q_\theta((x_e, x_i))\big)$
            \STATE Compute total loss: $\mathcal{L}_t = \mathcal{L}_{\text{student}} + \mathcal{L}_{\text{teacher}}$
            \STATE Update student parameters $\theta \leftarrow \theta - \eta \nabla_\theta \mathcal{L}_t$
        \ENDFOR

    \ENDIF
\ENDFOR
\RETURN Trained student model $\mathcal{S}_\theta$
\end{algorithmic}
\end{algorithm*}

\input{tables/digits_all_forgetting}
\input{tables/domainnet_all_forgetting_CR}

\begin{figure*}[h]
    \centering
    \includegraphics[width=0.8\textwidth]{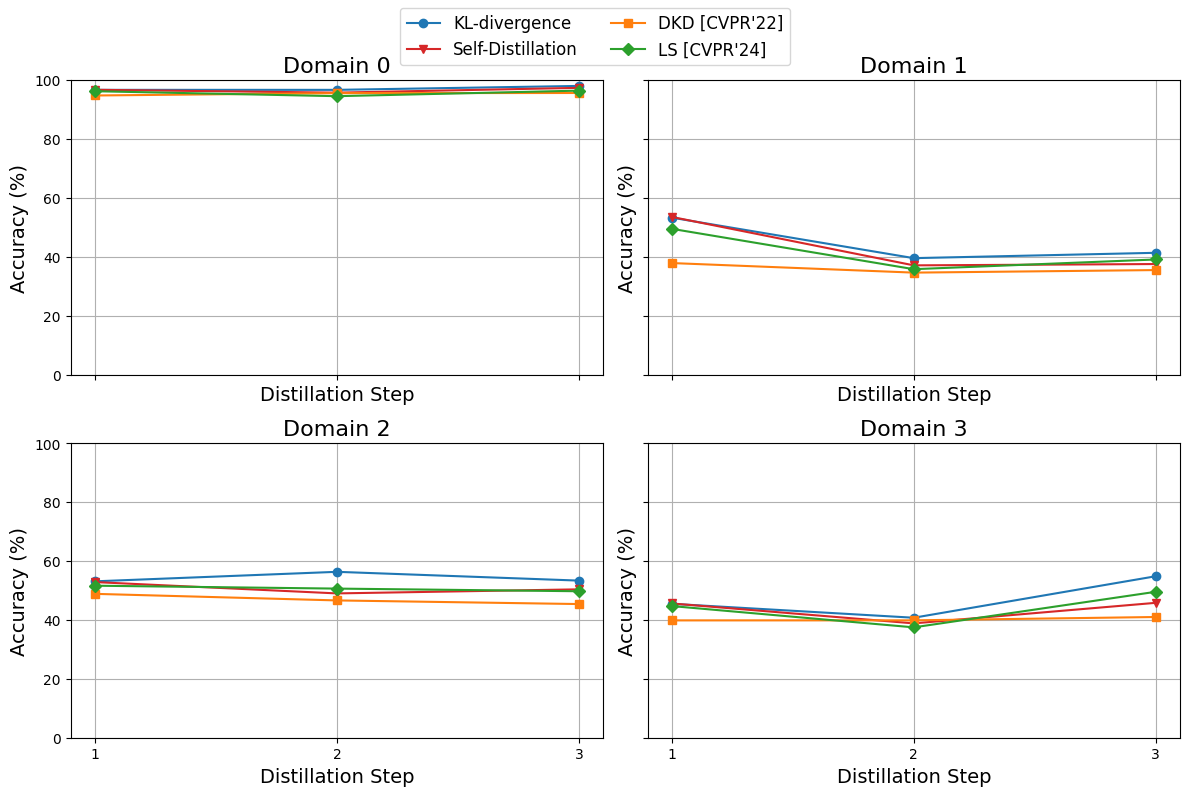}
    \caption{Accuracy of the student on all domains at all steps for the considered methods on CIFAR20, training with Internal Data only.}
    \label{fig:curve1}
\end{figure*}

\begin{figure*}[h]
    \centering
    \includegraphics[width=0.8\textwidth]{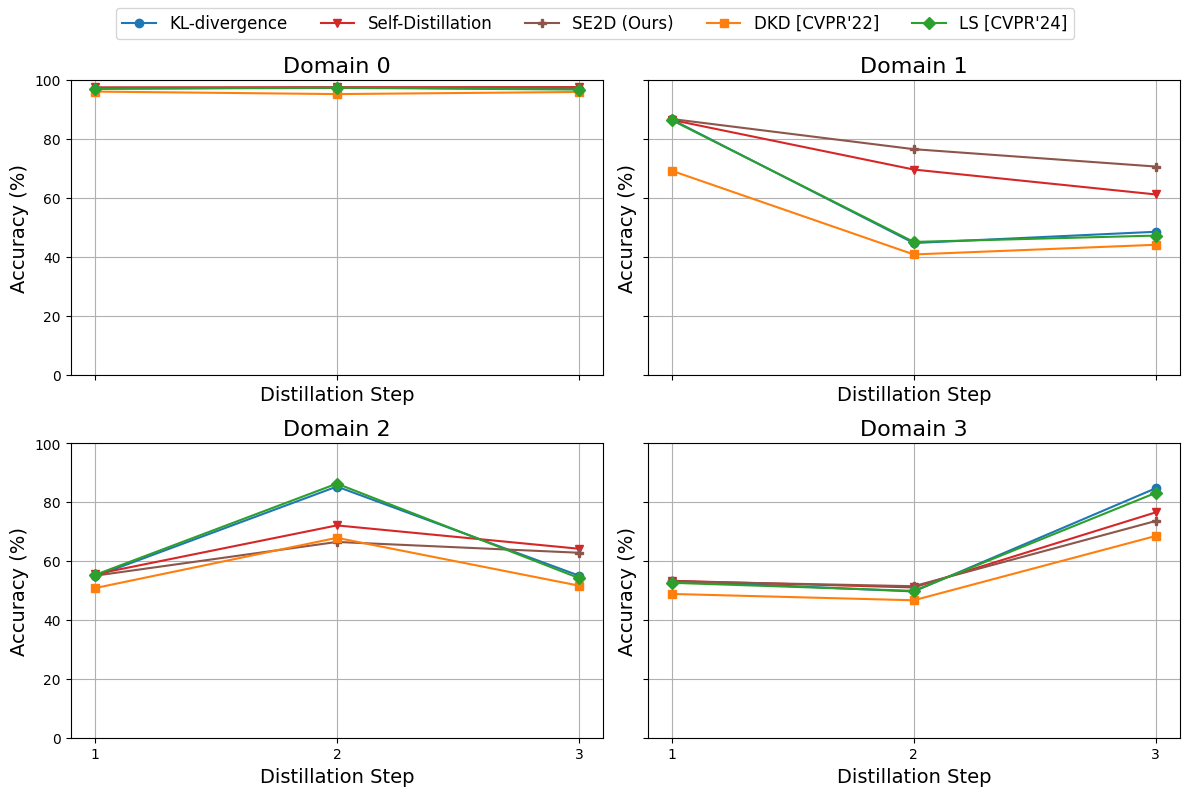}
    \caption{Accuracy of the student on all domains at all steps for the considered methods on CIFAR20, training with External Data.}
    \label{fig:curve2}
\end{figure*}

\begin{figure*}[h]
    \centering
    \includegraphics[width=0.8\textwidth]{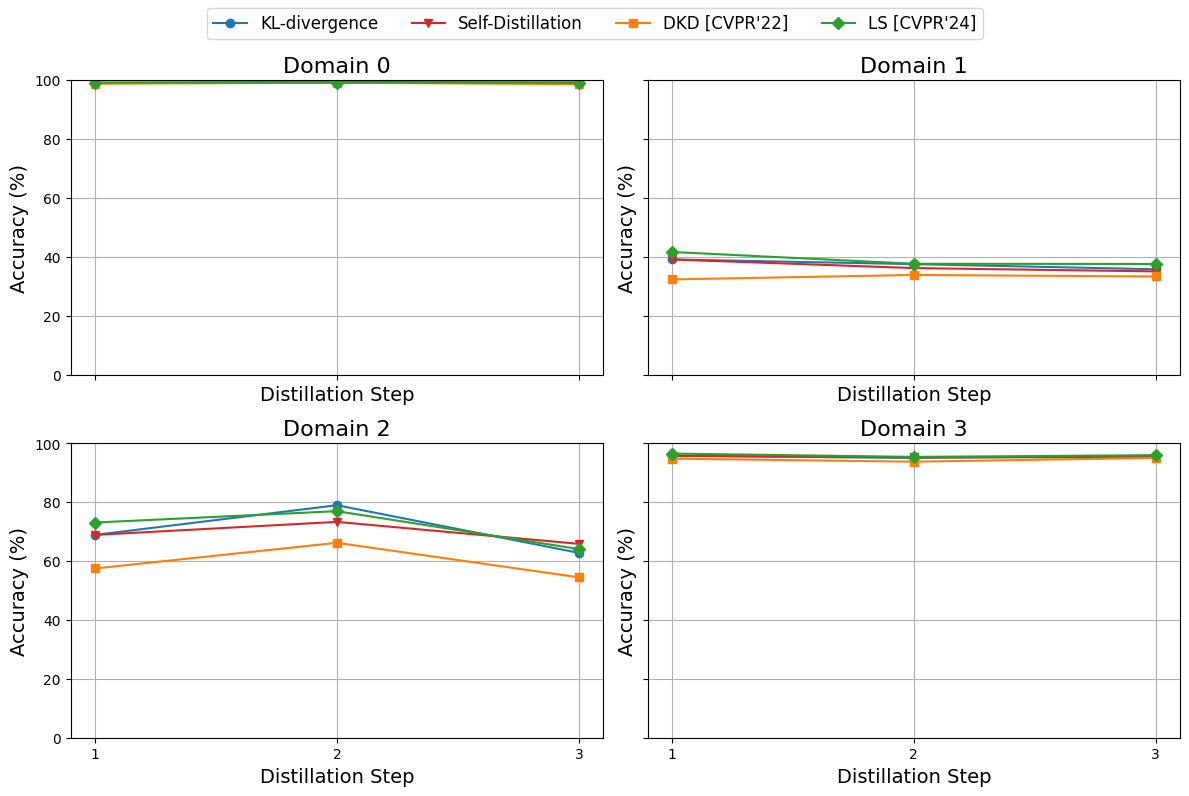}
    \caption{Accuracy of the student on all domains at all steps for the considered methods on Digits, training with Internal Data only.}
    \label{fig:curve3}
\end{figure*}

\begin{figure*}[h]
    \centering
    \includegraphics[width=0.8\textwidth]{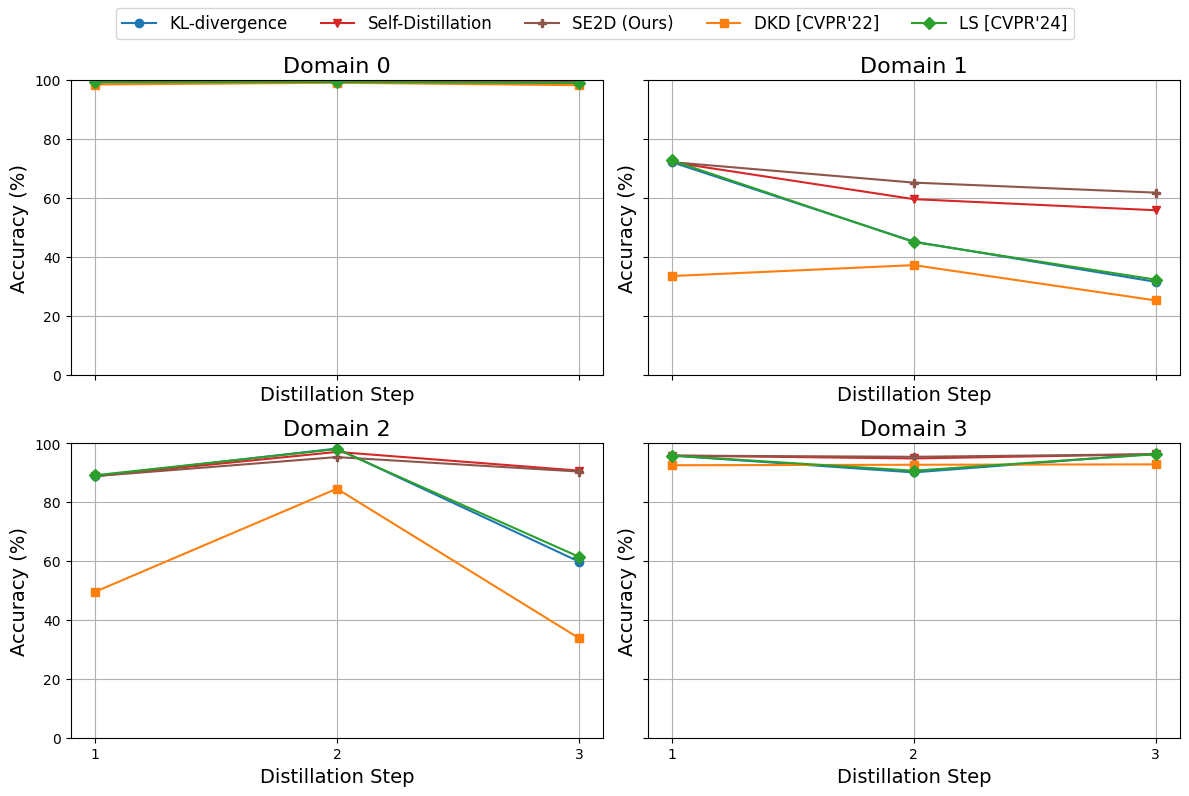}
    \caption{Accuracy of the student on all domains at all steps for the considered methods on Digits, training with External Data.}
    \label{fig:curve4}
\end{figure*}

\paragraph{Accuracy Curves}
We report per-domain accuracy curves during training with related external data in Figures~\ref{fig:curve1} to Figure~\ref{fig:curve4}. 

\subsection{Additional Architecures}
We report results with additional architectures. Namely, we experimented with a ViT-tiny as a student instead of the ViT-base version in the main manuscript. Similarly, we experimented with larger models as teacher using CLIP-base teachers (ViT-L/14). Results are presented in Table~\ref{tab:foundations}.
\input{tables/cifar_20_all_tiny}

\begin{table*}[h]
\centering
\renewcommand{\arraystretch}{1.2}
\setlength{\tabcolsep}{4pt}
\tiny

% --- Define colors ---
\definecolor{OutDomCol}{RGB}{255,178,178}
\definecolor{InDomCol}{RGB}{211,211,211}
\definecolor{DomainCol}{RGB}{255,255,255}
\definecolor{AvgCol}{RGB}{250,235,220}
\definecolor{OtherDom}{RGB}{230,240,250}

\caption{Domain Accuracy (\%, higher is better) on DomainNet with CLIP-based teachers. Mean and standard deviation are reported.}
\vspace{-2pt}

% ---------------- Block 1: Internal Data Only (Seq -1) ----------------
\resizebox{0.95\textwidth}{!}{%
\begin{tabular}{
    >{\columncolor{DomainCol}\color{black}}l
    >{\columncolor{InDomCol}}c
    *{4}{>{\columncolor{OtherDom}}c}
    >{\columncolor{DomainCol}}c
    >{\columncolor{AvgCol}}c
}
\toprule
 & \multicolumn{7}{c}{\textbf{DomainNet - Internal Data Only}} \\
\hline
\textbf{Method} & Clipart & Infograph & Painting & Quickdraw & Real & Sketch \xmark & Avg. \\
\hline
$\mathcal{T}_{best}$ (upper bound) & 86.33 & 53.89 & 79.74 & 69.62 & 89.86 & - & 76.23 \\
KL-divergence      & 78.33 $\pm$ 0.23 & \underline{19.00 $\pm$ 0.30} & 39.25 $\pm$ 0.81 & \underline{35.36 $\pm$ 0.49} & 51.51 $\pm$ 0.74 & - & 44.69 $\pm$ 0.32 \\
DKD [CVPR'22]      & \underline{78.55 $\pm$ 0.03} & 18.89 $\pm$ 0.41 & \underline{39.69 $\pm$ 0.19} & 34.77 $\pm$ 0.38 & \underline{52.45 $\pm$ 0.49} & - & \underline{44.87 $\pm$ 0.06} \\
LS [CVPR'24]       & 74.39 $\pm$ 0.39 & 15.54 $\pm$ 0.66 & 33.92 $\pm$ 0.56 & 20.09 $\pm$ 1.11 & 46.62 $\pm$ 0.76 & - & 38.11 $\pm$ 0.64 \\
MDS [ICLR'25]      & 75.51 $\pm$ 0.10 & 17.52 $\pm$ 0.20 & 35.67 $\pm$ 1.39 & 26.22 $\pm$ 0.70 & 47.71 $\pm$ 0.69 & - & 40.53 $\pm$ 0.27 \\
Self-Distillation  & \textbf{79.99 $\pm$ 0.55} & \textbf{21.72 $\pm$ 0.80} & \textbf{43.65 $\pm$ 0.57} & 26.52 $\pm$ 1.42 & \textbf{57.21 $\pm$ 0.13} & - & \textbf{45.82 $\pm$ 0.61} \\
\bottomrule
\end{tabular}
}

\vspace{10pt}

% ---------------- Block 2: Related External Data (Seq 0) ----------------
\resizebox{0.95\textwidth}{!}{%
\begin{tabular}{
    >{\columncolor{DomainCol}\color{black}}l
    >{\columncolor{InDomCol}}c
    *{4}{>{\columncolor{OtherDom}}c}
    >{\columncolor{OutDomCol}}c
    >{\columncolor{AvgCol}}c
}
\toprule
 & \multicolumn{7}{c}{\textbf{DomainNet + Related External Data}} \\
\hline
\textbf{Method} & Clipart & Infograph & Painting & Quickdraw & Real & Sketch \xmark & Avg. \\
\hline
$\mathcal{T}_{best}$ (upper bound) & 86.33 & 53.89 & 79.74 & 69.62 & 89.86 & - & 76.23 \\
KL-divergence      & 78.22 $\pm$ 0.39 & 19.98 $\pm$ 0.17 & 41.59 $\pm$ 0.55 & \textbf{42.83 $\pm$ 0.37} & 52.62 $\pm$ 0.40 & - & \underline{47.05 $\pm$ 0.22} \\
DKD [CVPR'22]      & 78.17 $\pm$ 0.11 & 20.01 $\pm$ 0.19 & 41.78 $\pm$ 0.39 & \underline{42.69 $\pm$ 0.84} & 52.37 $\pm$ 0.24 & - & 47.00 $\pm$ 0.01 \\
LS [CVPR'24]       & 74.64 $\pm$ 0.13 & 16.63 $\pm$ 0.31 & 37.09 $\pm$ 0.09 & 26.78 $\pm$ 0.35 & 47.45 $\pm$ 0.35 & - & 40.52 $\pm$ 0.17 \\
MDS [ICLR'25]      & 75.21 $\pm$ 0.59 & 18.55 $\pm$ 0.45 & 39.32 $\pm$ 0.13 & 31.84 $\pm$ 0.73 & 49.32 $\pm$ 0.39 & - & 42.85 $\pm$ 0.39 \\
Self-Distillation  & \textbf{79.47 $\pm$ 0.50} & \textbf{22.84 $\pm$ 0.60} & \textbf{47.51 $\pm$ 1.33} & 30.15 $\pm$ 0.77 & \textbf{58.51 $\pm$ 1.37} & - & \textbf{47.69 $\pm$ 0.84} \\
\hline
SE2D (ours)        & \underline{78.52 $\pm$ 0.10} & \underline{21.34 $\pm$ 0.45} & \underline{47.12 $\pm$ 0.35} & 29.18 $\pm$ 0.30 & \underline{58.01 $\pm$ 0.61} & - & 46.83 $\pm$ 0.31 \\
\bottomrule
\end{tabular}
}

\vspace{-4pt}
\label{tab:foundations}
\end{table*}

\subsection{Additional Sequences}
As presented in the main paper, DomainNet is particularly challenging. Therefore, we conducted additional experiments where challenging domains have either been removed or used as external data. Such results are presented in Table~\ref{tab:add_seq}. Despite SE2D's lower overall performance compared to Self-Distillation, the disparity between the two methods diminishes in less complex scenarios, where SE2D simultaneously expands its lead over the baseline.

\begin{table*}[h]
\centering
\renewcommand{\arraystretch}{1.2}
\setlength{\tabcolsep}{4pt}
\tiny
    
% --- Fixed Color Definitions ---
\definecolor{ClipartGrey}{RGB}{211,211,211}
\definecolor{UsedBlue}{RGB}{230,240,250}
\definecolor{EDRed}{RGB}{255,178,178}       % Added missing value/brace here
\definecolor{IgnoredWhite}{RGB}{255,255,255}
\definecolor{AvgCol}{RGB}{250,235,220}

% Mapping names to the commands used in your table structure:
\colorlet{InDomCol}{ClipartGrey}
\colorlet{OtherDom}{UsedBlue}
\colorlet{OutDomCol}{EDRed}
\colorlet{DomainCol}{IgnoredWhite}

\caption{Accuracy per domain (\%). \colorbox{ClipartGrey}{Grey}: Internal Data; \colorbox{UsedBlue}{Blue}: Domain known by the teacher (active); \colorbox{EDRed}{Red}: External Data (ED); \colorbox{IgnoredWhite}{White}: Ignored. Avg computed on Internal Data + Active domains.}
\vspace{-2pt}

% ---------------- Block 1: Sequence 1 ([2, 4, 5]) ----------------
\resizebox{0.95\textwidth}{!}{%
\begin{tabular}{
    l
    >{\columncolor{ClipartGrey}}c
    >{\columncolor{IgnoredWhite}}c
    >{\columncolor{UsedBlue}}c
    >{\columncolor{EDRed}}c
    >{\columncolor{UsedBlue}}c
    >{\columncolor{UsedBlue}}c
    >{\columncolor{AvgCol}}c
}
\toprule
 & \multicolumn{7}{c}{\textbf{DomainNet - Sequence 1 (Quickdraw is used as ED and Infograph is ignored)}} \\
\hline
\textbf{Method} & Clipart & Infograph & Painting & Quickdraw & Real & Sketch & Avg. \\
\hline
KL-divergence      & 74.66 $\pm$ 0.14 & - & 33.01 $\pm$ 0.21 & - & 46.17 $\pm$ 0.62 & 48.68 $\pm$ 0.15 & 50.63 \\
DKD [CVPR'22]      & 76.18 $\pm$ 0.20 & - & 36.20 $\pm$ 0.49 & - & 49.70 $\pm$ 0.62 & 54.85 $\pm$ 0.32 & 54.23 \\
MDS [ICLR'25]      & 70.76 $\pm$ 0.40 & - & 30.01 $\pm$ 0.78 & - & 43.52 $\pm$ 1.09 & 51.56 $\pm$ 0.68 & 48.96 \\
Self-Distillation  & \textbf{80.43 $\pm$ 0.10} & - & \textbf{47.47 $\pm$ 0.71} & - & \textbf{61.10 $\pm$ 0.59} & \underline{58.15 $\pm$ 0.31} & \textbf{61.79} \\
\hline
SE2D (ours)        & \underline{76.89 $\pm$ 0.25} & - & \underline{38.72 $\pm$ 0.49} & - & \underline{52.83 $\pm$ 0.20} & \textbf{58.84 $\pm$ 0.25} & \underline{56.82} \\
\bottomrule
\end{tabular}
}

\vspace{10pt}

% ---------------- Block 2: Sequence 2 ([2, 3, 4, 5]) ----------------
\resizebox{0.95\textwidth}{!}{%
\begin{tabular}{
    l
    >{\columncolor{ClipartGrey}}c
    >{\columncolor{EDRed}}c
    >{\columncolor{UsedBlue}}c
    >{\columncolor{UsedBlue}}c
    >{\columncolor{UsedBlue}}c
    >{\columncolor{UsedBlue}}c
    >{\columncolor{AvgCol}}c
}
\toprule
 & \multicolumn{7}{c}{\textbf{DomainNet - Sequence 2 (Infograph is used as ED)}} \\
\hline
\textbf{Method} & Clipart & Infograph & Painting & Quickdraw & Real & Sketch & Avg. \\
\hline
KL-divergence      & 75.78 $\pm$ 0.14 & - & 34.61 $\pm$ 0.52 & 16.59 $\pm$ 0.22 & 49.27 $\pm$ 0.43 & 52.43 $\pm$ 0.44 & 45.73 \\
DKD [CVPR'22]      & 76.73 $\pm$ 0.33 & - & 36.48 $\pm$ 0.68 & 17.53 $\pm$ 0.44 & 51.84 $\pm$ 0.55 & 56.83 $\pm$ 0.34 & 47.88 \\
MDS [ICLR'25]      & 75.28 $\pm$ 0.17 & - & 35.43 $\pm$ 0.92 & 16.47 $\pm$ 0.48 & 50.74 $\pm$ 0.88 & 56.65 $\pm$ 0.37 & 46.91 \\
Self-Distillation  & \textbf{80.45 $\pm$ 0.06} & - & \underline{48.48 $\pm$ 0.41} & \textbf{20.84 $\pm$ 0.89} & \textbf{64.79 $\pm$ 0.40} & \textbf{59.04 $\pm$ 0.24} & \textbf{54.72} \\
\hline
SE2D (ours)        & \underline{78.08 $\pm$ 0.20} & - & \textbf{49.02 $\pm$ 0.36} & \underline{18.50 $\pm$ 0.40} & \underline{63.29 $\pm$ 0.32} & \underline{58.99 $\pm$ 0.22} & \underline{53.58} \\
\bottomrule
\end{tabular}
}

\vspace{10pt}

% ---------------- Block 3: Sequence 3 ([2, 4]) ----------------
\resizebox{0.95\textwidth}{!}{%
\begin{tabular}{
    l
    >{\columncolor{ClipartGrey}}c
    >{\columncolor{IgnoredWhite}}c
    >{\columncolor{UsedBlue}}c
    >{\columncolor{IgnoredWhite}}c
    >{\columncolor{UsedBlue}}c
    >{\columncolor{EDRed}}c
    >{\columncolor{AvgCol}}c
}
\toprule
 & \multicolumn{7}{c}{\textbf{DomainNet - Sequence 3 (Sketch is used as ED, Infograph and Quickdraw are ignored)}} \\
\hline
\textbf{Method} & Clipart & Infograph & Painting & Quickdraw & Real & Sketch & Avg. \\
\hline
KL-divergence      & 75.75 $\pm$ 0.35 & - & 42.44 $\pm$ 0.67 & - & 65.15 $\pm$ 0.53 & - & 61.11 \\
DKD [CVPR'22]      & 76.25 $\pm$ 0.23 & - & 45.52 $\pm$ 0.72 & - & 70.50 $\pm$ 0.12 & - & 64.09 \\
MDS [ICLR'25]      & 75.25 $\pm$ 0.07 & - & 44.74 $\pm$ 0.94 & - & 70.11 $\pm$ 0.35 & - & 63.37 \\
Self-Distillation  & \textbf{79.54 $\pm$ 0.26} & - & \textbf{59.61 $\pm$ 0.30} & - & \textbf{71.48 $\pm$ 0.26} & - & \textbf{70.21} \\
\hline
SE2D (ours)        & \underline{78.04 $\pm$ 0.15} & - & \underline{59.36 $\pm$ 0.68} & - & \underline{70.61 $\pm$ 0.40} & - & \underline{69.34} \\
\bottomrule
\end{tabular}
}

\vspace{-4pt}
\label{tab:add_seq}
\end{table*}

% \subsection{Additional Datasets}
% We report additional results with external datasets, notably by mixing Domainnet with CUB and MNIST. The results are presented in Table~\ref{tab:domainnet_mixedd}.
% \input{tables/domainnet_mixedd}

\section{Algorithms}
To provide a clear overview of our training methodology, we present the Continual Distillation procedure in Algorithm~\ref{algo:cd}. Furthermore, a detailed description of our proposed SE2D approach is provided in Algorithm~\ref{alg:ours}.

%% file: tables/cifar_20_all_forgetting.tex
\begin{table}[t]
\large
\centering
\renewcommand{\arraystretch}{1}
\setlength{\tabcolsep}{5pt}

\definecolor{OutDomCol}{RGB}{255,178,178}
\definecolor{InDomCol}{RGB}{211,211,211}
\definecolor{DomainCol}{RGB}{255,255,255}
\definecolor{AvgCol}{RGB}{250,235,220}
\definecolor{OtherDom}{RGB}{230,240,250}

\caption{Forgetting (\%, lower is better) of the student at the end of training on CIFAR20 for 4 scenarios. Internal Data Only (D0), Related External Data (D4), CUB as ED, and MNIST as ED. The number of runs is set to $3$.}

% ---------------- Block 1 ----------------
\resizebox{0.48\textwidth}{!}{%
\begin{tabular}{
    >{\columncolor{DomainCol}\color{black}}l
    >{\columncolor{InDomCol}}c
    *{3}{>{\columncolor{OtherDom}\color{black}}c}
    >{\columncolor{DomainCol}}c
    >{\columncolor{AvgCol}}c
}
\toprule
\multicolumn{6}{c}{\textbf{CIFAR20 - Internal Data Only}} \\
\hline
\textbf{Method} & D0 & D1 & D2 & D3 & D4 \xmark & Avg. (0-3) \\
\hline
KL-divergence      & 0 & 11.95 & 2.98 & 0 & - & 3.73 \\
DKD [CVPR'22]      & 0 & \textbf{2.38} & 3.47 & 0 & - & \textbf{1.46} \\
LS [CVPR'24]       & 0 & 10.42 & \textbf{1.88} & 0 & - & 3.08 \\
MDS [ICLR'25]      & 0 & 11.95 & 2.98 & 0 & - & 3.23 \\
Self-Distillation  & 0 & 16.02 & 2.47 & 0 & - & 4.12 \\
\end{tabular}
}

% ---------------- Block 2 ----------------
\resizebox{0.48\textwidth}{!}{%
\begin{tabular}{
    >{\columncolor{DomainCol}\color{black}}l
    >{\columncolor{InDomCol}}c
    *{3}{>{\columncolor{OtherDom}\color{black}}c}
    >{\columncolor{OutDomCol}}c
    >{\columncolor{AvgCol}}c
}
\toprule
\multicolumn{6}{c}{\textbf{CIFAR20 + Related External Data}} \\
\hline
\textbf{Method} & D0 & D1 & D2 & D3 & D4 & Avg. (0-3) \\
\hline
KL-divergence      & 0.52 & 38.15 & 30.26 & 0 & - & 17.23 \\
DKD [CVPR'22]      & 0.13 & 25.15 & 16.25 & 0 & - & 10.38 \\
LS [CVPR'24]       & 0.62 & 39.23 & 32.08 & 0 & - & 17.98 \\
MDS [ICLR'25]      & 0.29 & 37.74 & 31.00 & 0 & - & 17.26 \\
Self-Distillation  & 0 & 25.33 & 7.93 & 0 & - & 8.32 \\
\hline
SE2D (ours)        & 0 & \textbf{16.10} & \textbf{3.67} & 0 & - & \textbf{4.44} \\
\end{tabular}
}

% ---------------- Block 3 ----------------
\resizebox{0.48\textwidth}{!}{%
\begin{tabular}{
    >{\columncolor{DomainCol}\color{black}}l
    >{\columncolor{InDomCol}}c
    *{3}{>{\columncolor{OtherDom}\color{black}}c}
    >{\columncolor{OutDomCol}}c
    >{\columncolor{AvgCol}}c
}
\toprule
\multicolumn{6}{c}{\textbf{CIFAR20 + CUB}} \\
\hline
\textbf{Method} & D0 & D1 & D2 & D3 & CUB & Avg. (0-3) \\
\hline
KL-divergence      & 0.93 & 30.68 & 17.82 & 0 & - & 12.36 \\
DKD [CVPR'22]      & 0.89 & \textbf{5.37} & 3.86 & 0 & - & \textbf{2.03} \\
LS [CVPR'24]       & 2.54 & 32.28 & 20.66 & 0 & - & 11.87 \\
MDS [ICLR'25]      & 0.61 & 30.79 & 16.26 & 0 & - & 11.41 \\
Self-Distillation  & 0.81 & 27.88 & \textbf{2.48} & 0 & - & 7.29 \\
\hline
SE2D (ours)        & \textbf{0.40} & 21.61 & 6.81 & 0 & - & 7.21 \\
\midrule
\multicolumn{6}{c}{\textbf{CIFAR20 + MNIST}} \\
\hline
\textbf{Method} & D0 & D1 & D2 & D3 & MNIST & Avg. (0-3) \\
\hline
KL-divergence      & 0 & 15.20 & 7.40 & 0 & - & 5.15 \\
DKD [CVPR'22]      & 0 & \textbf{2.74} & \textbf{0.57} & 0 & - & \textbf{0.83} \\
LS [CVPR'24]       & 1.57 & 14.58 & 9.50 & 0 & - & 5.41 \\
MDS [ICLR'25]      & 0.97 & 15.34 & 7.67 & 0 & - & 5.00 \\
Self-Distillation  & 0 & 9.69 & 1.02 & 0 & - & 2.18 \\
\hline
SE2D (ours)        & 0.85 & 9.88 & 2.85 & 0 & - & 4.43 \\
\bottomrule
\end{tabular}
}

\label{tab:fgt_c20}
\end{table}

%% file: tables/digits_all_forgetting.tex
\begin{table*}[h]
\centering
\tiny
\renewcommand{\arraystretch}{1}
\setlength{\tabcolsep}{5pt}

% --- Define colors ---
\definecolor{OutDomCol}{RGB}{255,178,178}   % light red
\definecolor{InDomCol}{RGB}{211,211,211}    % light gray
\definecolor{DomainCol}{RGB}{255,255,255}
\definecolor{AvgCol}{RGB}{250,235,220}      % light orange
\definecolor{OtherDom}{RGB}{230,240,250}    % light blue

% --- Table ---
\caption{Forgetting (\%, lower is better) of the student at the end of training on Digits for 2 scenarios. Internal Data Only (D0), Related External Data (D4). The number of runs is set to 3.}

% ---------------- Block 1 ----------------
\resizebox{0.76\textwidth}{!}{%
\begin{tabular}{
    >{\columncolor{DomainCol}\color{black}}l
    >{\columncolor{InDomCol}}c
    *{3}{>{\columncolor{OtherDom}\color{black}}c}
    >{\columncolor{DomainCol}}c
    >{\columncolor{AvgCol}}c
}
\toprule
\multicolumn{7}{c}{\textbf{Digits - Internal Data Only}} \\
\hline
\textbf{Method} & MNIST & SVHN & MNIST-M & USPS & KMNIST \xmark & Avg. \\
\hline
KL-divergence      & 0.23 & 3.34 & 16.20 & 0 & - & 4.94 \\
DKD [CVPR'22]      & 0.47 & \textbf{0.53} & 11.73 & 0 & - & \textbf{3.12} \\
LS [CVPR'24]       & \textbf{0.10} & 4.09 & 12.88 & 0 & - & 4.40 \\
MDS [ICLR'25]      & 0.23 & 3.34 & 16.20 & 0 & - & 4.94 \\
Self-Distillation  & 0.13 & 4.06 & \textbf{7.46} & 0 & - & 3.54 \\
\end{tabular}
}

\vspace{0.5em} % Adds a small gap between the two sub-tables

% ---------------- Block 2 ----------------
\resizebox{0.76\textwidth}{!}{%
\begin{tabular}{
    >{\columncolor{DomainCol}\color{black}}l
    >{\columncolor{InDomCol}}c
    *{3}{>{\columncolor{OtherDom}\color{black}}c}
    >{\columncolor{OutDomCol}}c
    >{\columncolor{AvgCol}}c
}
\toprule
\multicolumn{7}{c}{\textbf{Digits + Related External Data}} \\
\hline
\textbf{Method} & MNIST & SVHN & MNIST-M & USPS & KMNIST & Avg. \\
\hline
KL-divergence      & 0.24 & 40.68 & 38.41 & 0 & - & 19.17 \\
DKD [CVPR'22]      & 0.84 & 12.00 & 50.77 & 0 & - & 15.87 \\
LS [CVPR'24]       & 0.27 & 40.50 & 36.59 & 0 & - & 19.22 \\
MDS [ICLR'25]      & 0.25 & 40.69 & 38.35 & 0 & - & 19.16 \\
Self-Distillation  & \textbf{0.10} & 16.33 & 6.35 & 0 & - & 5.58 \\
\hline
SE2D (ours)        & 0.13 & \textbf{10.37} & \textbf{4.90} & 0 & - & \textbf{3.73} \\
\bottomrule
\end{tabular}
}

\label{tab:fgt_digit}
\end{table*}

%% file: tables/domainnet_all_forgetting_CR.tex
\begin{table*}[h]
\centering
\renewcommand{\arraystretch}{1}
\setlength{\tabcolsep}{5pt}
\tiny

% --- Define colors ---
\definecolor{OutDomCol}{RGB}{255,178,178}
\definecolor{InDomCol}{RGB}{211,211,211}
\definecolor{DomainCol}{RGB}{255,255,255}
\definecolor{AvgCol}{RGB}{250,235,220}
\definecolor{OtherDom}{RGB}{230,240,250}

% --- Table ---
\caption{Forgetting (\%, lower is better) on DomainNet for 2 scenarios: Internal Data Only (Seq -1) and Related External Data (Seq 0). Averaged across 3 runs.}
\vspace{-2pt}

% ---------------- Block 1: Seq -1 (Internal Data Only) ----------------
\resizebox{0.76\textwidth}{!}{%
\begin{tabular}{
    >{\columncolor{DomainCol}\color{black}}l
    >{\columncolor{InDomCol}}c
    *{4}{>{\columncolor{OtherDom}}c}
    >{\columncolor{DomainCol}}c
    >{\columncolor{AvgCol}}c
}
\toprule
 & \multicolumn{7}{c}{\textbf{DomainNet - Internal Data Only}} \\
\hline
\textbf{Method} & Clipart & Infograph & Painting & Quickdraw & Real & Sketch \xmark & Avg. \\
\hline
KL-divergence      & 4.63 & 9.91 & 13.70 & 20.31 & 0 & - & 12.14 \\
DKD [CVPR'22]      & 3.85 & 10.34 & 13.89 & 22.38 & 0 & - & 12.61 \\
LS [CVPR'24]       & 3.50 & 11.48 & 15.39 & 23.73 & 0 & - & 13.52 \\
MDS [ICLR'25]      & 5.20 & 10.70 & 15.69 & 21.46 & 0 & - & 13.26 \\
Checkpoint         & \textbf{1.27} & \textbf{5.86} & \textbf{11.30} & 20.68 & 0 & - & \textbf{9.78} \\
\hline
SE2D (ours)        & 4.60 & 9.58 & 13.49 & \textbf{20.27} & 0 & - & 11.98 \\
\end{tabular}
}

\vspace{5pt}

% ---------------- Block 2: Seq 0 (Related External Data) ----------------
\resizebox{0.76\textwidth}{!}{%
\begin{tabular}{
    >{\columncolor{DomainCol}\color{black}}l
    >{\columncolor{InDomCol}}c
    *{4}{>{\columncolor{OtherDom}}c}
    >{\columncolor{OutDomCol}}c
    >{\columncolor{AvgCol}}c
}
\toprule
 & \multicolumn{7}{c}{\textbf{DomainNet + Related External Data}} \\
\hline
\textbf{Method} & Clipart & Infograph & Painting & Quickdraw & Real & Sketch \xmark & Avg. \\
\hline
KL-divergence      & 5.56 & 11.89 & 18.07 & 29.12 & 0 & - & 16.16 \\
DKD [CVPR'22]      & 4.64 & 12.15 & 18.31 & 29.69 & 0 & - & 16.20 \\
LS [CVPR'24]       & 5.40 & 13.70 & 20.98 & 32.44 & 0 & - & 18.13 \\
MDS [ICLR'25]      & 6.25 & 13.03 & 19.68 & 29.62 & 0 & - & 17.15 \\
Checkpoint         & \textbf{1.70} & \textbf{6.63} & \textbf{14.54} & \textbf{29.43} & 0 & - & \textbf{13.08} \\
\hline
SE2D (ours)        & 3.42 & 6.91 & 15.54 & 30.86 & 0 & - & 14.18 \\
\bottomrule
\end{tabular}
}

\vspace{-4pt}
\label{tab:fgt_domainnet_cr}
\end{table*}

%% file: tables/cifar_20_all_tiny.tex
\begin{table*}[h]
\centering
\renewcommand{\arraystretch}{1.2} % Slightly increased for better readability with standard deviations
\setlength{\tabcolsep}{5pt}

% --- Define colors ---
\definecolor{OutDomCol}{RGB}{255,178,178}   % light red
\definecolor{InDomCol}{RGB}{211,211,211}    % light gray
\definecolor{DomainCol}{RGB}{255,255,255}   % white
\definecolor{AvgCol}{RGB}{250,235,220}      % light orange
\definecolor{OtherDom}{RGB}{230,240,250}    % light blue

% --- Table ---
\caption{Performances (\%, higher is better) of the student at the end of training on CIFAR20 for 2 scenarios with a ViT-tiny. Internal Data Only (D0), Related External Data (D4). The number of runs is set to 3. Average and standard deviations are reported.}

% ---------------- Block 1 ----------------
\resizebox{0.76\textwidth}{!}{%
\begin{tabular}{
    >{\columncolor{DomainCol}\color{black}}l
    >{\columncolor{InDomCol}}c
    *{3}{>{\columncolor{OtherDom}\color{black}}c}
    >{\columncolor{DomainCol}}c
    >{\columncolor{AvgCol}}c
}
\toprule
\multicolumn{7}{c}{\textbf{CIFAR20 - Internal Data Only}} \\
\hline
\textbf{Method} & D0 & D1 & D2 & D3 & D4 \xmark & Avg. (0-3) \\
\hline
$\mathcal{T}_{best}$ (upper bound) & 97.75 & 95.80 & 96.70 & 95.75 & - & 96.5 \\
KL-divergence & 96.27 $\pm$ 0.21 & 37.85 $\pm$ 0.78 & 52.28 $\pm$ 0.42 & 49.48 $\pm$ 1.28 & - & 58.97 $\pm$ 0.67 \\
DKD [CVPR'22] & 95.40 $\pm$ 1.05 & 33.27 $\pm$ 1.60 & 45.38 $\pm$ 0.90 & 40.45 $\pm$ 1.34 & - & 53.62 $\pm$ 1.22  \\
LS [CVPR'24]  & 96.25 $\pm$ 0.85 & \textbf{38.95 $\pm$ 2.17} & 51.58 $\pm$ 2.47 & \textbf{50.15 $\pm$ 4.40} & - & \textbf{59.23 $\pm$ 2.47}  \\
MDS [ICLR'25] & 94.45 $\pm$ 0.44 & 35.08 $\pm$ 0.88 & 45.35 $\pm$ 0.41 & 41.98 $\pm$ 3.10 & - & 54.22 $\pm$ 1.21 \\
Self-Distillation & \textbf{96.27 $\pm$ 0.35} & 37.93 $\pm$ 1.22 & \textbf{51.82 $\pm$ 0.53} & 46.43 $\pm$ 0.97 & - & 58.11 $\pm$ 0.77 \\
\bottomrule
\end{tabular}%
}

\vspace{1em}

% ---------------- Block 2 ----------------
\resizebox{0.76\textwidth}{!}{%
\begin{tabular}{
    >{\columncolor{DomainCol}\color{black}}l
    >{\columncolor{InDomCol}}c
    *{3}{>{\columncolor{OtherDom}\color{black}}c}
    >{\columncolor{OutDomCol}}c
    >{\columncolor{AvgCol}}c
}
\toprule
\multicolumn{7}{c}{\textbf{CIFAR20 + Related External Data}} \\
\hline
\textbf{Method}  & D0 & D1 & D2 & D3 & D4  & Avg. (0-3)\\
\hline
$\mathcal{T}_{best}$ (upper bound) & 97.75 & 95.80 & 96.70 & 95.75 & - & 96.5 \\
KL-divergence &   96.36 $\pm$ 0.33 & 46.54 $\pm$ 1.17 & 55.86 $\pm$ 1.10 & 75.96 $\pm$ 1.76 & - & 68.68 $\pm$ 1.09 \\
DKD [CVPR'22] &    95.18 $\pm$ 0.46 & 40.77 $\pm$ 0.95 & 50.92 $\pm$ 1.07 & 60.69 $\pm$ 1.32 & - & 61.89 $\pm$ 0.95 \\
LS [CVPR'24]  &  96.45 $\pm$ 0.26 & 46.80 $\pm$ 0.74 & 55.08 $\pm$ 0.61 & \textbf{76.49 $\pm$ 1.50} & - & 68.7 $\pm$ 0.78 \\
Self-Distillation &   \textbf{97.26 $\pm$ 0.25} & 55.42 $\pm$ 0.31 & \textbf{61.62 $\pm$ 0.70} & 68.74 $\pm$ 1.46 & - & 70.76 $\pm$ 0.68 \\
\hline
SE2D (ours) &   96.77 $\pm$ 0.25 & \textbf{62.33 $\pm$ 1.19} & 59.45 $\pm$ 1.36 & 65.82 $\pm$ 2.03 & - & \textbf{71.09 $\pm$ 1.21} \\
\bottomrule
\end{tabular}%
}
\label{tab:cifar_tiny}
\end{table*}